\documentclass[10pt,twocolumn,letterpaper]{article}

\usepackage[dvipsnames]{xcolor}
\usepackage{iccv}
\usepackage{times}
\usepackage{epsfig}
\usepackage{graphicx}
\usepackage{amsmath}
\usepackage{amssymb}

\usepackage[utf8]{inputenc} 
\usepackage[T1]{fontenc}    
\usepackage{url}            
\usepackage{booktabs}       
\usepackage{amsfonts}       
\usepackage{nicefrac}       
\usepackage{microtype}      
\usepackage{soul}
\usepackage{tabularx}
\usepackage{xspace}
\usepackage{multirow}
\usepackage{enumitem}
\usepackage{float}
\usepackage{setspace}
\usepackage{array}
\usepackage{wrapfig}
\usepackage{subcaption}
\usepackage{pifont}
\usepackage{colortbl}
\usepackage[font=small,labelfont=bf]{caption}

\usepackage[pagebackref=true,breaklinks=true,letterpaper=true,colorlinks,bookmarks=false]{hyperref}
\usepackage{url}

\usepackage[capitalize]{cleveref}
\crefname{section}{Sec.}{Secs.}
\Crefname{section}{Section}{Sections}
\Crefname{table}{Table}{Tables}
\crefname{table}{Tab.}{Tabs.}

\def\modelname{CLIPpy\xspace}
\def\datasetname{HQITP\xspace}

\definecolor{Gray}{gray}{0.90}

\newcommand{\JS}[1]{{\textcolor{DarkOrchid}{#1}}}

\newcommand{\citet}{\cite}
\newcommand{\citep}{\cite}

\newcommand{\cmark}{\ding{51}}%
\newcommand{\xmark}{\ding{55}}%

\newcommand{\inc}[1]{\ensuremath{_{\text{\textcolor{ForestGreen}{(+#1)}}}}}

\newlength\savewidth\newcommand\shline{\noalign{\global\savewidth\arrayrulewidth
  \global\arrayrulewidth 1pt}\hline\noalign{\global\arrayrulewidth\savewidth}}

\iccvfinalcopy 

\def\iccvPaperID{5241} 

\begin{document}

\title{Perceptual Grouping in Contrastive Vision-Language Models}

\author{
Kanchana Ranasinghe\thanks{Work performed as part of Apple internship.} \,, Brandon McKinzie, Sachin Ravi, \\
Yinfei Yang, Alexander Toshev, Jonathon Shlens\thanks{Work performed at Apple.} \\
Apple \\
{\tt\small kranasinghe@cs.stonybrook.edu}
}

\maketitle
\ificcvfinal\thispagestyle{empty}\fi

\begin{abstract}
\vspace{-0.5em}
Recent advances in zero-shot image recognition suggest that vision-language models learn generic visual representations with a high degree of semantic information that may be arbitrarily probed with natural language phrases. Understanding an image, however, is not just about understanding {\it what} content resides within an image, but importantly, {\it where} that content resides. In this work we examine how well vision-language models are able to understand where objects reside within an image and group together visually related parts of the imagery. We demonstrate how contemporary vision and language representation learning models based on contrastive losses and large web-based data capture limited object localization information. We propose a minimal set of modifications that results in models that uniquely learn both semantic and spatial information. We measure this performance in terms of zero-shot image recognition, unsupervised bottom-up and top-down semantic segmentations, as well as robustness analyses. We find that the resulting model achieves state-of-the-art results in terms of unsupervised segmentation, and demonstrate that the learned representations are uniquely robust to spurious correlations in datasets designed to probe the causal behavior of vision models.
\end{abstract}

\vspace{-1em}
\section{Introduction}
\vspace{-0.1em}

\begin{figure}[t]
\begin{center}
\includegraphics[width=0.48\textwidth]{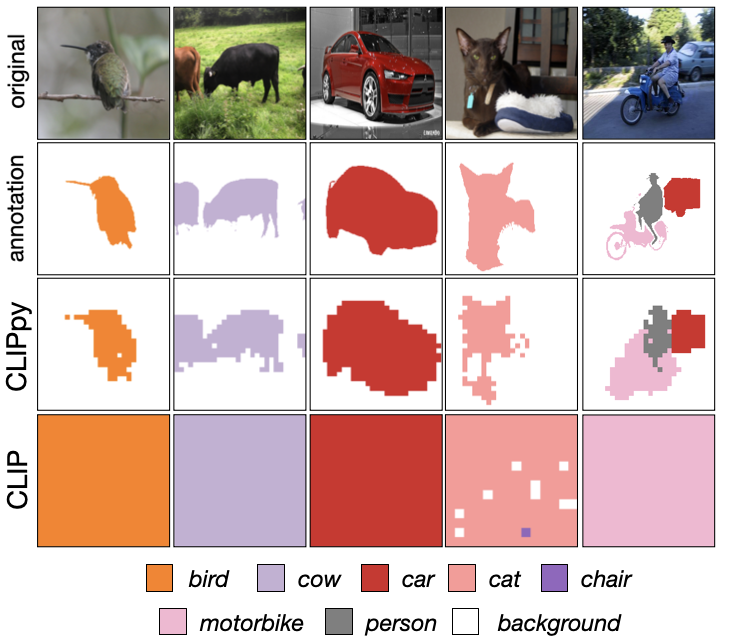}
\end{center}
\vspace{-1.7em}
\caption{\small
{\bf Semantic localization in contrastive VLMs.} We measure the ability of vision-language models to predict a label at each spatial position in a zero shot manner based on the similarity of location tokens to the corresponding language tokens on selected examples. CLIP / ALIGN \cite{jia2021scaling,radford2021clip} have minimal understanding of the spatial location of individual objects (row 4). Our proposed \modelname (row 3) predicts the label at locations that correspond closely to human annotation for semantic segmentation (row 2). All predictions were performed with no access to any segmentation data during training or inference. More visualizations in App. \ref{section:appendix-qualitative}.}
\label{fig:motivation}
\vspace{-1.4em}
\end{figure}

Learning a representation for visual imagery requires resolving not only what resides within an image, but also where that information resides \cite{marr1982vision}. In many applications, knowledge of {\it where} information resides is sometimes more important than a precise description of the content \cite{geiger2012we,sun2020scalability}. Hence, our ability to learn more generic and robust visual representations requires learning the geometry of visual semantics, and how visual information may be grounded by specific regions of the visual field.

While recent vision-language models trained under weak supervision demonstrate a remarkable ability to learn generic and transferable visual representations \cite{jia2021scaling,radford2021clip,yu2022coca,desai2021virtex}, they showcase a profound inability to associate visual content with individual objects (Fig. \ref{fig:motivation}, bottom row). In other words, models trained on large weakly-supervised data have a limited ability to group together visually related content \cite{ghiasi2022open}. Because the representations have a poor understanding of {\it where} an object resides, they easily conflate background with foreground content.
Hence, the learned representations are unable to learn the spatial layout of a scene \cite{subramanian2022reclip,thrush2022winoground}, and are susceptible to learning spurious correlations between a semantic label and extraneous content \cite{sagawa2019distributionally,liu2021just}.

Recent work \cite{xu2022groupvit, Xu2023LearningOS} attempts to bridge this gap through grouping mechanisms under the same weakly supervised training paradigm, but focus more on foreground objects (neglecting background classes). Another direction is task specific unsupervised fine-tuning \cite{Zhou2021ExtractFD, Ding2021DecouplingZS} which loses the generic and transferable nature of these representations. 

In this work, we explore vision-language models that learn from similar weakly labeled data, but a) retain the generic and transferable nature of features, and b) learns where all (background and foreground) visual content resides within an image. 
Unlike previous attempts using grouping specific architectures \cite{xu2022groupvit, Xu2023LearningOS} or dense human annotations \cite{ghiasi2022open, gu2021vild, li2022language}, we explore a minimal set of modifications to existing CLIP models \cite{radford2021clip} that leads to grouping of visual imagery while retaining their weakly supervised and scalable training procedure. 
We find that two small adjustments -- employing specific pretraining strategies and adjusting spatial feature aggregation 
-- results in models that are equally effective in zero-shot image recognition, but also retain spatial information regarding object locations (see Fig. \ref{fig:motivation}, 3rd row). 

The resulting model termed \modelname exhibits \textit{perceptual grouping} -- that is, the ability to select and combine related visual signals into semantically meaningful regions  \cite{wertheimer1923,marr1982vision,roelfsema2006cortical}. Endowing models with perceptual grouping -- whether in a bottom up (based solely on visual content) or top down (guided by external information, language in this case) manner -- in learned representations has been a long standing goal in computer vision \cite{malik2001visual,malik2016three}. In this work, our key contributions are as follows:
\setlist{nolistsep}
    \begin{itemize}[noitemsep,leftmargin=*]
\item Identify systematic failure of contrastive vision-language models \cite{radford2021clip, jia2021scaling} to properly identify where objects reside within an image, and group semantically related content.
\item Design a minimal set of changes to endow these model with perceptual grouping, resulting in state-of-the-art zero-shot segmentation \textit{without} training on \textit{any} segmentation data or performing task specific fine-tuning.
\item Emergence of localization ability in our models uniquely leads to robustness to counterfactual manipulations. The degree of robustness matches if not surpasses previous state-of-the-art supervised learning methods employing specialized training methodologies.
\end{itemize}


\section{Related Work}

{\bf Vision-language models for grounding.}
Contrastive language image pre-training \cite{radford2021clip} (CLIP) led to a range of follow up work performing open-vocabulary detection \cite{gu2021vild, kamath2021mdetr, Li2022AdaptingCF, Li2021GroundedLP, Zeng2021MultiGrainedVL,Dou2022CoarsetoFineVP} or segmentation \cite{ghiasi2022open, li2022language, Zhang2022GLIPv2UL}. While these methods leverage dense human annotations for training, an alternate line of works \cite{xu2022groupvit, Xu2023LearningOS, Zhou2021ExtractFD, yao2022filip,cui2022democratizing} attempt 
to learn alignment between regions of images and language
with only image level noisy captions for supervision. Their weak supervision allows better scalability (to more data) leading to learning more generic and transferable representations.  
In fact, multiple such works \cite{xu2022groupvit, Xu2023LearningOS, Zhou2021ExtractFD, Ding2021DecouplingZS, li2022language} perform zero-shot
semantic segmentation. However, unlike \cite{xu2022groupvit, Xu2023LearningOS} geared to segment a fixed count of foreground objects, our proposed \modelname can better segment arbitrary object counts and background classes. In contrast to \cite{Zhou2021ExtractFD} using generic image level features, \modelname explicitly learns local features during training. Moreover, \modelname requires no dense human annotations or task-specific fine-tuning in contrast to \cite{Ding2021DecouplingZS,li2022language}. We also highlight how \cite{xu2022groupvit, Xu2023LearningOS, Ding2021DecouplingZS} perform grouping independent of language at inference - however \modelname can group conditioned on language, capturing variable object boundaries for different language prompts. 

Multiple \textbf{contemporary works} also explore similar directions as \modelname, leveraging pre-trained vision-language models for various grouping tasks under weak supervision (no pixel level annotation) \cite{Zhang2023AssociatingSG,Luo2022SegCLIPPA,Cha2022LearningTG,Mukhoti2022OpenVS,Burgert2022PeekabooTT,Karazija2023DiffusionMF}. Combining self-supervised methods that emerge grouping \cite{caron2021emerging} with CLIP models \cite{radford2021clip} for cross-modal alignment is explored in \cite{Zhang2023AssociatingSG} gaining notable improvements at object boundaries. A clustering mechanism containing learnable centres similar to \cite{xu2022groupvit} is combined with reconstruction and super-pixel alignment losses to achieve grouping in \cite{Luo2022SegCLIPPA}. Learning decoder networks over a frozen CLIP backbone \cite{radford2021clip} with text to image patch similarity losses are explored in \cite{Cha2022LearningTG,Mukhoti2022OpenVS} resulting in similar grouping behaviour. In contrast to these methods utilizing contrastive vision language training to emerge grouping, recent works \cite{Burgert2022PeekabooTT,Karazija2023DiffusionMF} also showcase how text-to-image generative models (particularly Stable Diffusion \cite{Rombach2021HighResolutionIS}) can be leveraged to perform visual grouping.

\begin{table}[t]
    \centering
    \scalebox{0.83}{
    \def\arraystretch{1.0}
    \setlength\tabcolsep{1.0em}
    \begin{tabular}{l|c|c|c}
    \toprule
    Component      		& CLIP \cite{radford2021clip} & CLIP$^\dagger$ &  CLIPpy \\ \midrule
    Image Backbone      & ViT-B/16 & ViT-B/16    & ViT-B/16 \\
    Text Backbone       & T-B      & T-5         & T-5      \\
    Image Init 			& Random   & Random      & \JS{DINO}     \\
    Text Init 			& Random   & Random      & \JS{Sent T-5} \\	
    Image Pooling		& CLS      & CLS  	 	 & \JS{Max}      \\	
    Text Pooling		& Avg      & Avg  	 	 & Avg      \\
    Dataset		        & 300M$^*$ & CC-12M      & CC-12M  	\\
    VOC mIoU (\%)	    & 16.4     & 17.5        & \textbf{50.8} \inc{33.3}  \\
    VOC JS (\%)	        & 28.6     & 37.3        & \textbf{47.5} \inc{10.2}  \\
    \bottomrule
    \end{tabular}}
    \vspace{-0.8em}
    \caption{\small
    We \JS{highlight} the minimal differences of CLIPy from CLIP.
    CLIP$^\dagger$ is our implementation following train settings identical to CLIPpy. $^*$indicates OpenAI private data.
    }
    \label{tbl:compare}
    \vspace{-1.5em}
\end{table}
\begin{figure*}[t]
\begin{center}
\includegraphics[width=0.78\textwidth]{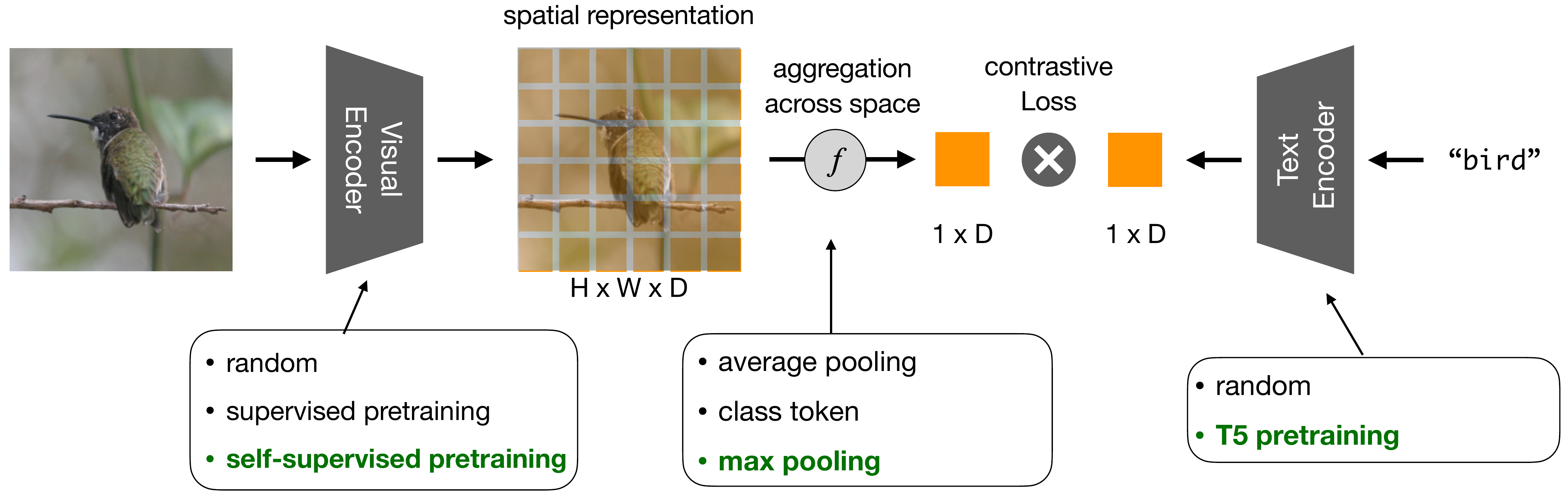}
\end{center}
\vspace{-1.6em}
\caption{\small
{\bf Architecture diagram.} Images and captions are separately embedded into Euclidean spaces, where image features are spatially aggregated. A contrastive loss trains the aggregated image embedding to be close to the caption embedding. We demonstrate that two minimal design decisions (indicated in green) are of paramount importance for CLIP \cite{radford2021clip} models to perform perceptual grouping under image-level weak supervision.}
\label{fig:diagram}
\vspace{-0.9em}
\end{figure*}

{\bf Zero-shot semantic segmentation.}
A form of top-down grouping, this relatively new task  \cite{zhao2017openvocabulary_scene_parsing,ji2018end,xian2019semantic,bucher2019zeroshotseg,STRICT,hu2020uncertainty, li2020consistent, baek2021exploiting, shen2021conterfactual} attempts to segment unseen classes, usually after a supervised training phase often involving dense annotation based supervision. 
Following two early representative works \cite{xian2019semantic, bucher2019zeroshotseg}, most later approaches \cite{li2020consistent,CaGNet,CaGNetv2,kendall2017uncertainties,shen2021conterfactual,cap2seg} formulate the task as a pixel-level zero-shot classification problem with a closed set vocabulary.
While \modelname follows a similar pixel based formulation, in contrast, our method requires no dense human annotations for supervision, no task specific fine-tuning, and is open-vocabulary.  
Recent work \cite{Ding2021DecouplingZS, Li2022AdaptingCF} also explores region-level classification leveraging pre-trained CLIP models \cite{radford2021clip}, but unlike \modelname perform grouping independent of language during inference.

{\bf Unsupervised segmentation}.
Analogous to bottom-up grouping, these works perform class-agnostic segmentation within the visual modality with no explicit language alignment \cite{caron2021emerging, hamilton2022unsupervised, VanGansbeke2021UnsupervisedSS, Ji2019InvariantIC,MelasKyriazi2022DeepSM}. This topic has a long, rich history in human visual perception \cite{wertheimer1923} and computer vision  \cite{malik2001visual}, and has been explored as means of generalizing to new visual domains \cite{qi2021open,malik2016three}. It is this goal that most closely inspires our work. 
Early efforts group pixels based on known spatially-local affinities \cite{comaniciu1997robust,shi2000normalized,ren2003learning}, with subsequent methods leading to region proposal networks for object detection \cite{uijlings2013selective} and advances in semantic segmentation \cite{arbelaez2012semantic}. Recent methods employ self-supervision to learn perceptual grouping  \cite{cho2021picie,hamilton2022unsupervised} or object-centric groupings \cite{Elsayed2022SAViTE, Locatello2020ObjectCentricLW, Wen2022SelfSupervisedVR, Bao2022DiscoveringOT, Henaff2022ObjectDA}.
Our proposed \modelname demonstrates competitive performance, but additionally aligns groups to the language modality explicitly.

{\bf Learning robust visual representations}. 
For a long time, ImageNet \cite{deng2009imagenet} accuracy was believed to provide a reasonable proxy for quality of learned visual representations \cite{girshick2014rcnn,kornblith2019better}.
However, recent work highlights notable deficiencies in such learned representations \cite{geirhos2021partial,recht2019imagenet,koh2021wilds} including sensitivity to low level textures, failure for domain shifts, and reliance on spurious correlations.
These failures inspired a large literature to mitigate learning spurious correlations \cite{sagawa2019distributionally,liu2021just,arjovsky2019invariant} by focusing on new optimization techniques. Progress on this issue may address parallel issues in fairness \cite{creager2021environment}. Resulting methods have largely focused on synthetic data, re-balancing data, and shaping learned embeddings \cite{nam2020learning,liu2021just}. Nonetheless, theoretical results suggest pessimistic bounds unless additional structure informs the problem (see refs. in \citet{sagawa2019distributionally}). Therein, the structured output predictions of proposed \modelname provide another promising solution. 
\section{Methodology}\label{sec:methods}

We first set the stage by discussing established core architectures and the contrastive learning formulation. Next, we discuss modifications that are the focus of the analysis in this work. In particular, we discuss aggregation options, pre-training alternatives, and token sub-sampling.

\subsection{Architecture and Training}
We provide a quick overview of our architecture  (Fig. \ref{fig:diagram}). Consider a batch size $N$, spatial height $H$, spatial width $W$, and depth $D$. $X$ is a tensor that has a shape of $[N, H, W, D]$ and is the output of an image encoder. $Y$ is a tensor that is of shape $[N, D]$ and is the output of a text encoder.

{\bf Language Model}. We employ a strong language model baseline derived from the transformer architecture \cite{vaswani2017attention} and implemented in T5 \cite{raffel2020exploring}. T5 models use an encoder-decoder architecture that is trained using a generative span corruption task, and have achieved state-of-the-art on a broad range of NLP tasks including GLUE~\cite{wang2018glue} and Super-Glue~\cite{wang2019superglue}. We use the encoder only and discard the decoder part. We employ the T5-base which consists of 12 transformer layers, 12 attention heads, and 768 token channel dimensions.

{\bf Image Model}. 
We explore two architectures for image featurization, CNN-based and Vision-Transformers, although we focus the majority of work on the latter. First, we employ the EfficientNet architecture \cite{tan2019efficientnet} as a high performant CNN architecture, which has been used previously in vision-language models. The specifics of the meta-architecture were derived from considerations based on neural architecture search.
Second, we employ the Vision Transformer (ViT) architecture \cite{dosovitskiy2020image}.
We refer the reader to \cite{dosovitskiy2020image,vaswani2017attention} for details. Briefly, ViT is largely inherited from the NLP literature and consists of a hierarchical associative memory. Each layer, termed a transformer, is composed of a Multi-headed Self-Attention (MSA) layer followed by a 2-layer feed-forward multi-layer perceptron (MLP).
The primary parameter of ViT is the patch size $P$ specifying the $P\times P$ patch of pixels constituting a token in the architecture.



{\bf Contrastive Representation Learning.} Let $x_i$ and $y_i$ denote the image and text embeddings (post aggregation) of the $i$'th example in the batch. A contrastive loss may be specified as the cross entropy across a batch \cite{radford2021clip,jia2021scaling}. The cross entropy is calculated between a one-hot encoding specifying the correspondence between the image and text examples, and a softmax-normalized distribution specifying the dot-product similarity between image and text embeddings.

\vspace{0.2em}
\scalebox{0.95}{
\hspace*{-2em}
\scriptsize
$L = \underbrace{-\frac{1}{N}\sum_{i=1}^N\log{\frac{\exp(x_i^\top y_i / \tau)}{\sum_{j=1}^{N} \exp(x_i^\top y_j / \tau)}}}_\text{image-to-text} \,\,+\,\, \underbrace{-\frac{1}{N}\sum_{i=1}^N\log{\frac{\exp(y_i^\top x_i / \tau)}{\sum_{j=1}^{N} \exp(y_i^\top x_j / \tau)}}}_\text{text-to-image}$
}
\vspace{0.2em}

The normalization for the image-to-text and text-to-image similarity is computed by summing over the potential matches (indexed by $j$) to the text and image examples within a batch, respectively. 
Note that $\tau$ is the temperature of the softmax for the normalization.

\subsection{Aggregation}
The goal of the aggregation method is to collapse the image embedding from a $[H, W, D]$ tensor to a $D$ dimensional vector.
\textbf{Average pooling} across space is an established technique for ensuring that the final embedding is independent of the image resolution \cite{szegedy2015going,long2015fully}, and has been adopted for CNN-based architectures in vision-language models \cite{jia2021scaling}. Alternatively, \textbf{maximum pooling} has been explored, in particular with success for point clouds~\cite{qi2017pointnet} and image-audio~\cite{Harwath2019JointlyDV}. Another approach typical for ViT borrowed from language modeling \cite{devlin2018bert} is the \textbf{class token (CLS)}, which is prepended to the image patch tokens \cite{dosovitskiy2020image}. A class token learns an embedding that aggregates information across all patch tokens in order to predict the image label. The class token may be used to summarize the content for an entire image for ViT-based models \cite{radford2021clip,caron2021emerging}. Subsequent work in vision-language models has explored learning pooling strategies  \cite{chen2021learning,yao2022filip}, heuristically selecting a set of similar neighbors  \cite{Yun_2022_CVPR} or learning attention-based mechanisms \cite{yu2022coca}.

In this work we systematically explore these aggregation strategies. In early experiments we found that many complex strategies for aggregation yielded poor results (App. \ref{section:additional-aggregation}). We found that the application of max pooling across the spatial dimensions -- while extremely simple -- was also by far the most effective (Sec. \ref{section:ablations}). We hypothesize that the success of max pooling may be due to the gradient updates being focused solely on a single spatial location, and not spread across all spatial dimensions.

\textcolor{black}{
\textbf{Why Max Pooling?}
In particular, the max pooling operation allows pre-aggregation features (shaped $[N, H, W, D]$) to determine the spatial location for gradient updates at each step, conditioned on input images. 
Across different images containing a common object at different spatial locations, the model has to select a conservative and minimal set of spatial locations for gradient updates.  
At the same time, given the cross-modal contrastive train objective, the aggregated feature of each such image must be aligned towards a common language concept (i.e. related to the common object).
We hypothesize that gradient updates at the common object's spatial location is the simplest optimization for the train objective in this case, leading to observed perceptual grouping.
}

\subsection{Pretraining}
{\bf Language Model}. For better sentence level representation, we utilize pre-training from Sentence-T5 \cite{ni2021sentence} which adapts a T5 encoder to sentence level embedding using a contrastive objective. We select Sentence-T5 over auto-regressive models such as \cite{devlin2018bert, Brown2020LanguageMA} because this contrastive loss is aligned to our setup. The model is 
trained on Stanford Natural Language Inference (SNLI) dataset with 275K examples focused on entailment questions \citep{bowman2015large,gao2021simcse}.

{\bf Image Model}. We investigate initializing the image model with several methods. First, we investigate initializing the image model using \textbf{supervised pre-training} and removing the final layer for logistic regression \cite{girshick2014rcnn,kornblith2019better}. We next investigate \textbf{self-supervised methods} derived from self-distillation (e.g. \cite{caron2021emerging}). We focused on this latter direction because such models demonstrated impressive performance in terms of localization \cite{caron2021emerging,hamilton2022unsupervised}. All image pre-training is performed on ImageNet-1K \cite{deng2009imagenet} dataset.

\vspace{0.5em}
\textbf{Suitable Visual Pre-training.}
The visual encoder representation space can be viewed as containing per-image features (post-aggregation) vs per-spatial location features (pre-aggregation). We hypothesize that semantics tied boundaries of this representation space should operate at the latter granularity to induce perceptual grouping. Furthermore, we suggest that initializations facilitating the former will detriment grouping behaviour. In particular, visual pre-training strategies separating image-level representations by semantics (e.g. supervised ImageNet pre-training) will diminish perceptual grouping. Self-supervised pre-training strategies focused on more granular within image representations (e.g. \cite{caron2021emerging}) will tend to enhance perceptual grouping. This hypothesis is empirically validated in ablations (see \Cref{tbl:ablation2}).



\subsection{Visual Token Sub-Sampling}
Motivated by vision transformers' ability to process sequences of length different to train time, we generate higher resolution segmentations during inference by sampling more image patches. In order to increase robustness to such varying resolution, we
utilize up to $2 \times$ higher resolution images during training but randomly drop 80\% of visual tokens to minimize additional compute overhead (similar to \cite{He2021MaskedAA,li2022scaling}). While improving segmentations, this also provides training stability possibly due to its regularizing effect (see App. \ref{section:training-details}).


\vspace{0.3em}
\subsection{Inference}
\label{subsec:inference}
\modelname performs inference under 3 different settings: a) classification, b) bottom-up grouping, and c) top-down grouping. 
On the visual modality, the first utilizes a spatially aggregated single per-image token while the latter two utilize sets of per-region tokens. 
Classification follows zero-shot analyses from \cite{radford2021clip} where the model is prompted at inference for a selection of labels (App. \ref{section:prompts} for prompts).
Bottom-up grouping follows a form of spectral clustering inspired by \cite{caron2021emerging} (refer to their demo). PCA on image features (from visual encoder pre-aggregation) gives top n(=8) principal components, which are used as cluster centers. Each of those same image features are assigned to one of the n clusters based on proximity (cosine similarity) to the centers, resulting in n clusters (or groups). 
Top-down grouping employs zero shot analysis similar to \cite{radford2021clip}, but at each spatial location, using the per-region tokens. This is similar to \cite{ghiasi2021simple} and generates predictions across space exploiting the transitive property of our aggregation operations. 
\section{Experiments}
\label{section:experiment-overview}

{\bf Experimental Setup.} We train our models on two datasets: Conceptual Captions 12M (CC-12M) \citep{changpinyo2021conceptual} and High Quality Image Text Pairs (\datasetname-134M) consisting of 12 million and 134 million image-text pairs, respectively (App. \ref{section:internal-dataset} for details).
For both datasets, text is  tokenized, and images resized and center cropped to 224$\times$224 pixels. We report results on EfficientNet-B5 employed by ALIGN \cite{jia2021scaling}, and ViT-B/16 employed by CLIP \cite{radford2021clip} although we focus more on the latter.
We train models on 32 GPUs across 4 machines with PyTorch \cite{NEURIPS2019_bdbca288}. See App. \ref{section:training-details} for more details.
We evaluate across image classification, localization, and robustness tasks. For image classification, we employ the validation splits of ImageNet \citep{deng2009imagenet} and ImageNet-v2 \citep{recht2019imagenet}, and for robustness we employ the test split of Waterbirds \cite{sagawa2019distributionally}. These datasets contain 1000, 1000, and 3 classes respectively.
For segmentation tasks, we employ the validation splits of PASCAL VOC \citep{everingham2010pascal}, ADE20K \citep{zhou2018ade,cheng2021maskformer}, COCO \citep{coco}, COCO (Obj) \citep{coco}, and Cityscapes \cite{Cordts2016TheCD}. Each of these datasets contain 20, 150, 133, 80, and 27 labels, respectively. 

{\bf Baselines for comparison.}
Given that most competitive baselines are trained on private datasets, we first attempt to reproduce results by training models on a corpus of image-text pairs. In more detail, we train on the public CC-12M dataset \cite{changpinyo2021conceptual} to provide reproducible numbers and observe competitive performance given our data limitations. We also train on the larger \datasetname-134M dataset to verify scalability. 

We first measure the performance of CLIP \cite{radford2021clip} and ALIGN \cite{jia2021scaling} on zero-shot image classification on ImageNet and ImageNet-v2. 
\Cref{table:image-recognition} highlights these results. We take this as a starting point for subsequent work.
In the following experiments we attempt to address the following questions:\setlist{nolistsep}
    \begin{itemize}[noitemsep]
    \item What are the limitations of current vision-language models? (Fig. \ref{fig:motivation})
    \item Do we observe perceptual grouping in vision language models? 
    (\cref{table:bottom-up,table:more-unsupervised-segmentation,table:semantic-segmentation}).
    \item How resilient are vision-language models to counterfactual manipulations? (Fig. \ref{figure:robustness}).
    \item How important are each of the proposed model modifications? (\cref{tbl:ablation1,tbl:ablation2,tbl:ablation3,tbl:ablation4}).
\end{itemize}


\subsection{Limitations of vision-language models}

Visual representations learned in vision-language models exhibit an impressive ability to generalize across tasks \cite{radford2021clip,jia2021scaling}. However they also exhibit a profound shortcoming -- learned visual representations maintain minimal information about {\it where} an object resides, failing to properly recognize what parts of an image constitute an object.

Fig. \ref{fig:motivation} (bottom row) showcases failure of a CLIP model; namely, the model improperly conflates visual content not associated with an object with the actual object. This can be observed by measuring the similarity of each embedding at each spatial location with a label set using the method in \cite{ghiasi2021simple} (\cref{subsec:inference}). One consistently observes that the central object of interest is incorrectly predicted to reside at every spatial location. For instance, in the left example, the CLIP model predicts that a {\tt bird} resides at every spatial location.
In a CNN architecture, where spatial information is inherently preserved, we observe some improvement, but the larger issue of poor localization remains (see App. \ref{section:cnn-analysis} for details).

This failure of vision-language models to properly understand the spatial organization of information is consistent with earlier observations.
Ablation experiments in ViT models demonstrated that removing positional embeddings minimally detriments predictive performance \cite{dosovitskiy2020image,naseer2021intriguing,pmlr-v162-zhai22a,subramanian2022reclip}.
Without positional information, ViT models effectively learn representations as a ``bag of image patches'', ignoring the spatial organization.

In contrast, if we perform the same analysis on \modelname, we see that the model retains significant information about spatial information (Fig. \ref{fig:motivation}, 3rd row). We take these visualizations as an impetus for further investigation. In particular, we start by quantifying the ability of the model to arbitrarily group together semantically related pixels, and compare this to previous works.

\begin{table}[t]%
\small
\centering
\def\arraystretch{1.2}
\begin{tabular}{c|c|cc}
\toprule
  & Dataset & IN & IN-v2 \\
   \shline
   ALIGN \cite{jia2021scaling} &  ALIGN-1800M & 76.4 & 70.1 \\
   CLIP \cite{radford2021clip} &  CLIP-400M & 65.5 & 60.8 \\
   \hline
   CLIP $^\dagger$ &  CC-12M & 46.0 & 40.3 \\
   GroupViT \cite{xu2022groupvit} & CC-12M+YFCC & 42.9 & - \\
   GroupViT $^\dagger$  & CC-12M & 25.6 & 23.8 \\
   \rowcolor{Gray}  
   \modelname & CC-12M & 45.3 & 40.0 \\
   \hline
   ALIGN $^\dagger$ &  \datasetname-134M & 51.1 & 45.6 \\
   CLIP $^\dagger$ &  \datasetname-134M & 61.4 & 56.4 \\
   \rowcolor{Gray}  
   \modelname & \datasetname-134M & 60.3 & 54.8 \\
   \bottomrule
\end{tabular}
\vspace{-0.5em}
\caption{\small 
\textbf{\modelname achieves competitive zero-shot image recognition.} IN and IN-v2 denote ImageNet and ImageNet-v2 accuracy, respectively. $^\dagger$ indicates our implementation. \citet{jia2021scaling} evaluated at 640$\times$640; others evaluated at 224$\times$224. \modelname shows $\pm 0.5$ and $\pm 0.9$ IN acc. (5 runs) on {\footnotesize CC-12M} and {\footnotesize\datasetname-134M}, respectively.}
\label{table:image-recognition} 
\end{table}
\begin{figure}[t]
\centering
\begin{minipage}{\linewidth}
    \scalebox{1.0}{
    \begin{minipage}{\textwidth}
    \includegraphics[width=0.20\textwidth]{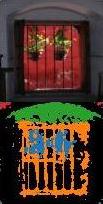} \hspace{-0.4em}
    \includegraphics[width=0.20\textwidth]{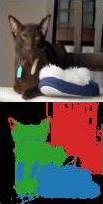} \hspace{-0.4em} 
    \includegraphics[width=0.20\textwidth]{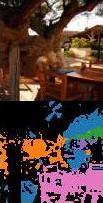} \hspace{-0.4em}
    \includegraphics[width=0.20\textwidth]{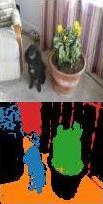} \hspace{-0.4em} 
    \includegraphics[width=0.20\textwidth]{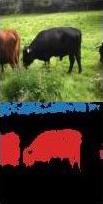}
    \end{minipage}
    }
\end{minipage}
\vspace{-0.5em}
\caption{\small
\textbf{Visualizations of bottom-up grouping by \modelname}. Each color represents one grouping learned on a given image.
}
 \label{figure:jaccard-similarity}
 \vspace{-1em}
\end{figure}

\subsection{Emergence of Bottom-Up Perceptual Grouping}

Unsupervised segmentation performance is a direct measure of bottom up perceptual grouping.  We apply \modelname at test time to perform semantic segmentation without prompting it for any labels 
\footnote{We perform PCA clustering (see \cref{subsec:inference}). GroupViT \cite{xu2022groupvit} \& DINO \cite{caron2021emerging} employ 8 \& 6 feature vectors based on their model architectures. Our visualizations employ 8 feature vectors (cluster centers).}. 
Fig. \ref{figure:jaccard-similarity} shows how the model visually groups semantically related regions of an image (see also Fig. \ref{fig:qualitative-bottom-up-segmentation-examples} in App.) as the image embeddings naturally group into spatially distinct clusters mirroring the image structure. We emphasize that this analysis does \textit{not} rely on text prompts \textit{nor} segmentation labels, but merely emerges from the image features alone. Hence the model has learned to \textit{group} perceptually related pixels merely based on the pixel content and associated image-level captions during training.

We quantify the accuracy of this bottom-up segmentation to capture known segmentations within annotated images. Following evaluation protocol in \citet{caron2021emerging, xu2022groupvit}, we compute the Jaccard Similarity (JS). JS here measures the average intersection over the union across all segmentation instances regardless of object category.
Our results in \cref{table:bottom-up,table:unsupervised-segmentation,table:more-unsupervised-segmentation} demonstrate competitive performance by \modelname. 
In VOC, \modelname achieves $54.6\%$ outperforming all previous models; in comparison, CLIP achieves 38.9\%. Additionally, on two more challenging datasets we note how the model drops in performance relatively, perhaps indicative of more visually cluttered scenes (Tab. \ref{table:more-unsupervised-segmentation}).
Our intuition for \modelname improving over CLIP is that CLS and average pooling breaks spatial structure of features, mixing image-level features across features at all spatial locations. 
We take these results to indicate that \modelname perceptually groups semantically related content better than previous work, providing state-of-the-art results in unsupervised segmentation.

\begin{table}[t]%
%
\small
\centering
\def\arraystretch{1.1}
\scalebox{0.92}{
\begin{tabular}{c|ccc|l}
\toprule
& Dataset & Train & SSP & VOC \\
\shline
DeiT \citet{caron2021emerging}  & \multirow{5}{*}{\small{ImageNet}} & class & \xmark & 24.6  \\
MoCo \citet{xu2022groupvit}     &  & self & \cmark &  28.2 \\
DINO \citet{caron2021emerging}  &  & self & \cmark &  45.9 \\
DSM \cite{MelasKyriazi2022DeepSM}& & self & \cmark & 37.2 \\
COMUS \cite{Zadaianchuk2022UnsupervisedSS} & & self & \cmark & 47.3 \\
\hline
DINO \citet{xu2022groupvit}  & \multirow{3}{*}{\shortstack[c]{\small{CC-12M \&} \\ \small{YFCC-100M}}} & self & \cmark & 41.8 \\
CLIP \citet{xu2022groupvit}  &    & text & \xmark & 28.6 \\
GroupViT \citet{xu2022groupvit} & & text & \xmark & 51.8   \\
\hline
CLIP $^\dagger$ & \multirow{3}{*}{\small{CC-12M}} & text & \xmark &  37.3 \\
GroupViT $^\dagger$ &   & text & \cmark & 42.8 \inc{5.5}  \\
\modelname & & text & \cmark & \textbf{47.5} \inc{10.2} \\ \hline
CLIP $^\dagger$ & \multirow{2}{*}{\small{\datasetname-134M}} & text & \xmark & 38.9 \\
\modelname & & text & \cmark & \textbf{54.6} \inc{15.7} \\
\bottomrule
\end{tabular}
}
\vspace{-0.7em}
\caption{\small
\textbf{\modelname effectively performs bottom-up grouping.} We report Jaccard Similarity, an instance average of IoU between predicted and annotated segmentations, independent of object labels. $^\dagger$denotes our implementations. SSP indicates the use of self-supervised visual pre-training.}
\label{table:bottom-up}
\vspace{-1em}
\end{table}

\begin{table}[t]
\begin{minipage}{\linewidth}
\begin{minipage}{0.68\linewidth}
    \centering
    \scalebox{0.9}{
        \small
        \centering
        \setlength\tabcolsep{0.5em}
        \begin{tabular}{c|c|ll}
        \toprule
        & Dataset  & ADE20K & COCO \\
        \shline
        CLIP $^\dagger$ & \multirow{2}{*}{\small{CC-12M}}  & 22.9 & 20.4 \\
        \modelname &  & \textbf{28.9} \inc{6.0} & \textbf{26.0} \inc{5.6} \\ \hline
        CLIP $^\dagger$ & \multirow{2}{*}{\scriptsize{\datasetname-134M}}  & 24.2 & 21.6 \\
        \modelname &  & \textbf{29.5} \inc{5.3} & \textbf{27.2} \inc{5.6} \\
        \bottomrule
        \end{tabular}
    }
\end{minipage}
\hspace{0.01\linewidth}
\begin{minipage}{0.28\linewidth}
\caption{\footnotesize
{\bf More bottom-up grouping:} \modelname improves Jaccard Similarity across datasets.
}
\label{table:more-unsupervised-segmentation}
\vspace{-1.0em}
\end{minipage}
\vspace{-0.5em}
\end{minipage}
\end{table}
\begin{table}[t]
\begin{minipage}{\linewidth}
\begin{minipage}{0.45\linewidth}
\scalebox{0.8}{
\centering
\def\arraystretch{0.85}
\setlength\tabcolsep{1.2em}
\begin{tabular}{l|c}
\toprule
Method                                          & JS   \\ \midrule
IIC \cite{Ji2019InvariantIC}     & 6.4  \\
MDC \cite{Caron2018DeepCF}    & 7.1  \\
PiCIE \cite{cho2021picie}        & 12.3 \\
STEGO \cite{hamilton2022unsupervised}  & 21.0 \\ \midrule
\modelname (ours)                & \textbf{22.3} \\ \bottomrule
\end{tabular}}
\end{minipage}
\hspace{0.01\linewidth}
\begin{minipage}{0.50\linewidth}
\caption{\small
{\bf More bottom-up grouping:} \modelname achieves competitive Jaccard Similarity (JS) on the Cityscapes Dataset 27 class segmentation setup \cite{hamilton2022unsupervised}. 
}
\label{table:unsupervised-segmentation}
\vspace{-1.0em}
\end{minipage}
\vspace{-0.5em}
\end{minipage}
\end{table}

\begin{table*}[t]
\small
\centering
\scalebox{0.96}{
\def\arraystretch{1.1}
\setlength\tabcolsep{1.0em}
\begin{tabular}{c|ccc|ll|ll}
    \toprule
   & Arch & Dataset & SSP & \small{ADE20K} & \small{COCO} & \small{VOC} & \small{COCO (obj)} \\
   \shline
    GroupViT \citet{xu2022groupvit} & ViT & \multirow{5}{*}{\small{CC-12M}} & \xmark & - & - & 41.1 & - \\
    GroupViT $^\dagger$             & ViT & & \cmark & 6.2 & 12.7 & 40.1 & 17.5 \\
    MaskCLIP $^\dagger$ \cite{Zhou2021ExtractFD} & ViT & & \xmark & 6.8 & 8.1 & 22.1 & 13.8 \\
    OVS \citet{Xu2023LearningOS}    & ViT & & \cmark & 7.1 & - & 44.6 & 25.1 \\
    \;\;CLIP $^\dagger$             & ViT & & \xmark & 5.0 & 7.8 & 17.5 & 13.2 \\
    \modelname                      & ViT & & \cmark & \textbf{13.1} \inc{8.1} & \textbf{23.8} \inc{16.0} & \textbf{50.8} \inc{33.3} & \textbf{28.5} \inc{15.3} \\
    \hline
    ALIGN \citet{ghiasi2021simple} & CNN & \small{ALIGN-1800M} & \xmark & 9.7 & 15.6 & - & - \\
    CLIP \citet{radford2021clip}  & ViT & \small{CLIP-400M}    & \xmark & 5.8 & 8.7 & 16.4 & 14.5 \\
    \hline
    \;\;\;ALIGN $^\dagger$ & CNN & \multirow{3}{*}{\small{\datasetname-134M}} & \xmark & 7.5 & 14.4 & 29.7 & - \\
    \;\;CLIP $^\dagger$ & ViT &  & \xmark & 5.1 & 8.0 & 18.1 & 14.5 \\
    \modelname & ViT &  & \cmark & \textbf{13.5} \inc{8.4} & \textbf{25.5} \inc{17.5} & \textbf{52.2} \inc{34.1} & \textbf{32.0} \inc{17.5} \\
   \bottomrule
\end{tabular}}
\vspace{-0.5em}
\caption{ \textbf{\modelname provides competitive localization with no segmentation or location annotations.} All models trained without any segmentation annotations. Results grouped by training dataset (bold highlights best per dataset). Numbers are mean IoU. $^\dagger$ indicates our implementation. SSP indicates image self-supervised pre-training to visual encoder.
}
\label{table:semantic-segmentation}
\vspace{-1.0em}
\end{table*}

\subsection{Top-down Grouping}

We demonstrated that \modelname is able to perceptually group visual content within an image.
Next, we ask how well this grouping corresponds to semantically meaningful labels. 
To measure the emergence of top-down grouping, we ask how well the perceptual grouping of the model may be steered by embeddings from the language model.
We test this 
by comparing the model's ability to perform zero-shot semantic segmentation across four datasets. Note that all of our results and comparisons are solely restricted to models trained on \textit{no} segmentation annotations\footnote{In App. \ref{section:prior-zemantic-segmentation}, we provide a summary of other zero-shot semantic segmentation results. Some of these prior results achieve superior performance, but we note that all of these methods were trained explicitly on various forms of segmentation masks, if not segmentation labels, often with task specific fine-tuning in contrast to the generic \& unsupervised nature of CLIPpy. 
}.

Fig. \ref{fig:motivation} provides a visualization of the predicted zero-shot segmentations (see also App. \ref{section:appendix-qualitative}), and Tab. \ref{table:semantic-segmentation} quantifies the results using mean intersection over union (mIoU). 
\modelname outperforms all other approaches on semantic segmentation when trained on the same datasets, both for CC-12M and HQITP-134M.
We view our datasets in two categories: ADE20K and COCO contain numerous background classes while VOC and  COCO (obj) contain only foreground object classes. We particularly highlight the notable performance improvement of \modelname for the former datasets.  
Moreover, in comparison to CLIP and ALIGN baselines, \modelname achieves significant improvements. We also 
replicate these baselines on the largest possible dataset within our compute budget (\datasetname), for comparison on a common dataset. These results on \datasetname also indicate clear performance improvements from \modelname. 

\begin{figure*}[t]%
\begin{minipage}{0.32\linewidth}
    \centering
    \includegraphics[width=1.0\linewidth]{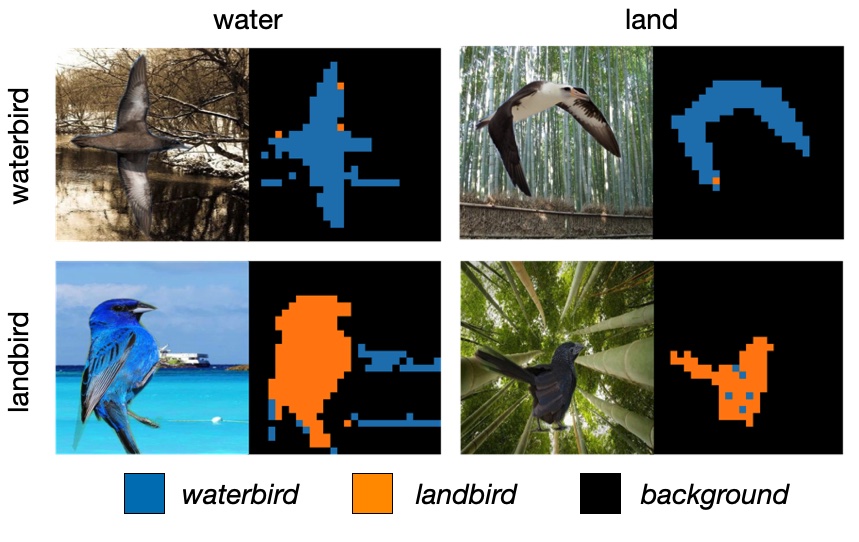}
\end{minipage}
\begin{minipage}{0.32\linewidth}
    \centering
    \def\arraystretch{1.3}
    \scriptsize
    \begin{tabular}{c|cc|c}
        \toprule
       CLIP & water & land & $\Delta$ \\
       \shline
       waterbird & 80.2 & 48.1 & {\bf \textcolor{red}{-32.1}}\\
       landbird & 38.8 & 71.7 & {\bf \textcolor{red}{-32.9}} \\
       \bottomrule
    \end{tabular} 
    \vspace{0.1cm} \\
    \begin{tabular}{c|cc|c}
        \toprule
       \modelname & water & land & $\Delta$ \\
       \shline
       waterbird & 76.9 & 74.9 & {\bf \textcolor{red}{-2.0}} \\
       landbird & 80.0 & 84.1 & {\bf \textcolor{red}{-4.1}}  \\
       \bottomrule
    \end{tabular} \\
\end{minipage}
\begin{minipage}{0.32\linewidth}
    \includegraphics[width=1.0\textwidth]{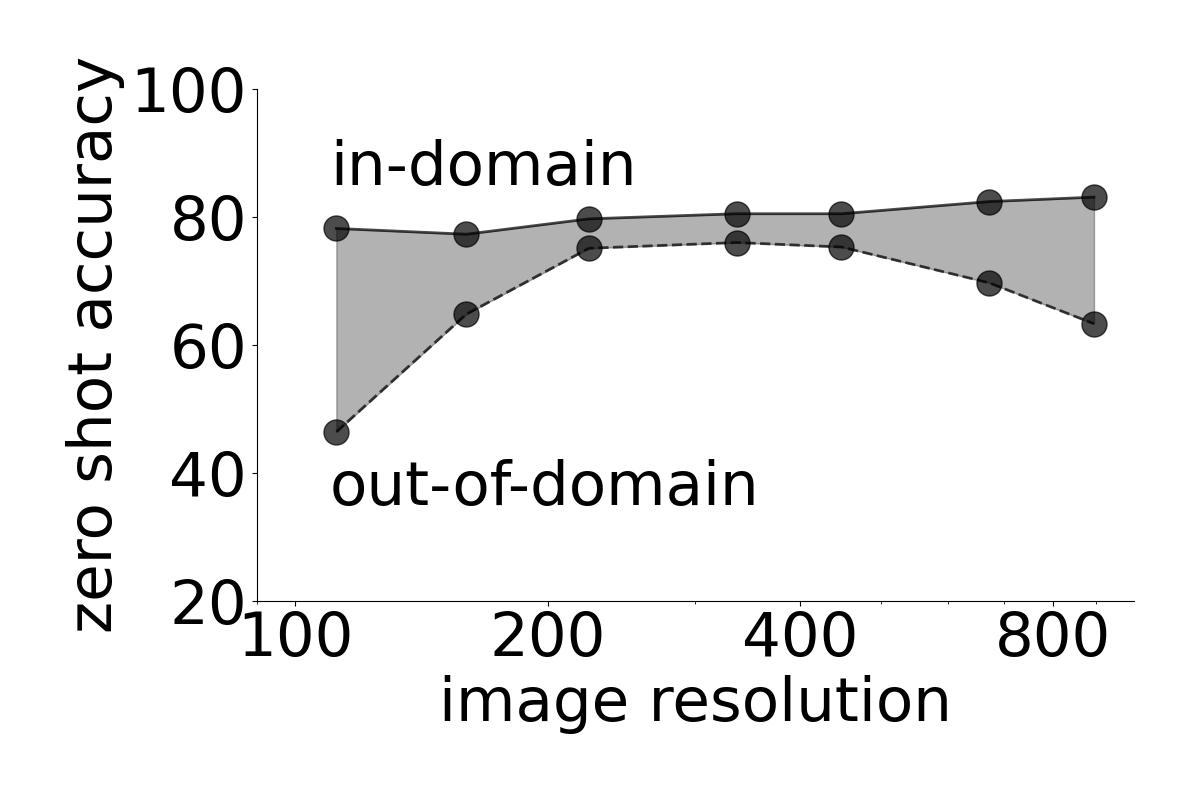}
\end{minipage}
\vspace{-1.0em}
\caption{\small 
\textbf{Perceptual grouping mitigates sensitivity to spurious correlations.} (left) Selected segmentation examples by \modelname of waterbirds and landbirds on each background. 
(centre) Accuracy on the {\it test} split (5794 examples) of Waterbirds on CLIP and \modelname evaluated at 448$\times$448 resolution. The domain gap $\Delta$ reports the drop in accuracy between on and off diagonal entries within a row. 
(right) Zero shot accuracy of \modelname across image resolution for landbirds on land (top) and water (bottom). Note log axis. Shading highlights $\Delta$.}
\label{figure:robustness}
\end{figure*}

GroupViT \cite{xu2022groupvit} and OVS \cite{Xu2023LearningOS} provide important points of comparison. These models use custom ViT architectures specific to grouping, are trained on common datasets (containing image-text pairs), and are designed to perform perceptual grouping by optimizing discretized attention masks. We draw attention to the clear performance improvements of \modelname over these methods across all datasets. We also highlight that OVS \cite{Xu2023LearningOS} uses pre-training strategies similar to ours. Our implementation of GroupViT also utilizes similar pre-training following \cite{Xu2023LearningOS}.
We take these results to mean that our simple changes to existing vision-language models uncover powerful localization information\footnote{We note that even removing all pretraining and solely training on CC-12M still retains notable grouping performance (\cref{tbl:ablation2}).}.

\subsection{Perceptual grouping may improve robustness}
\label{sec:robustness}

We have observed how parsimonious changes to vision-language models result in state-of-the-art unsupervised and zero-shot semantic segmentation. In this section, we ask how the resulting perceptual grouping may be exploited to improve the robustness of image understanding. 
A large literature has consistently observed that models systematically underperform under domain shifts \cite{recht2019imagenet}. For instance, CLIP, ALIGN, and \modelname underperform on ImageNet-v2 versus ImageNet (Tab. \ref{table:image-recognition}). Another means of assessing robustness is to measure how well a model \textit{causally} predicts the label from the appropriate input variates \cite{pearl2009causality,pearl2018book}. To probe for causal dependencies, one can measure model performance to counterfactual examples where an input is selectively manipulated in order to test for sensitivity to spurious correlations.  

A common formulation for this problem is to artificially synthesize a malicious dataset where a trained model may correlate inappropriate image features to predict a label \cite{xiao2020noise,moayeri2022comprehensive,jacobsen2018excessive,arjovsky2019invariant}. A large class of supervised learning algorithms have been developed to train on these datasets\footnote{Synthetic datasets are deliberately constructed to contain a class imbalance such that a minority class may be particularly prone to systematic worse performance. Consequently, experimenters have focused on the worse-case performance on the minority class \cite{sagawa2019distributionally,liu2021just}. Our work is instead focused on the domain gap to target the degree to which spurious correlations inappropriately influence predictions.}
with the aim of mitigating such spurious correlations \cite{sagawa2019distributionally,liu2021just,nam2020learning}.
One common synthetic benchmark is \textit{Waterbirds} \cite{sagawa2019distributionally} which places segmentations of birds in front of a background of land or water. The goal of any prediction system is a two-way classification of whether or not a bird is from the \texttt{waterbird} or \texttt{landbird} category. What makes this problem particularly challenging is when the background is not commensurate with the type of bird. For instance, a trained model may be prone to predict the type of bird due to the presence of water in the background in lieu of the visual appearance of the actual bird.

We first asked how our baseline CLIP model performs on this task when presented with a zero-shot three-way classification task (App. \ref{section:waterbirds-details} for inference procedure). Model performance depends heavily on the background (Fig. \ref{figure:robustness} centre). For instance, the prediction accuracy of \texttt{waterbirds} drops by $\Delta$=32.1\% (80.2 $\rightarrow$ 48.1) in the presence of an incommensurate background. Clearly, the baseline CLIP model performs zero-shot prediction by relying on features from the background. We note that open-source CLIP \cite{radford2021clip} has similar trends (see App. \cref{tbl:clip-waterbirds}). 

We next asked how \modelname performs given that it exhibits a unique ability to discriminate the spatial locations of objects. Fig. \ref{figure:robustness} shows selected examples from each class colored by the prediction at each spatial location. Clearly, the model is able to discriminate which locations correspond to each category. We quantify model accuracy across each task, and find the model far less sensitive to the background. 
For instance, in the case of waterbirds, \modelname  accuracy, while slightly less than the baseline CLIP model, only drops by $\Delta$ = 2.0\% (76.9 $\rightarrow$ 74.9) in spite of the background change (Fig. \ref{figure:robustness} right). 
Interestingly, the domain gap $\Delta$ is minimal ($\sim$4\%) around a broad range of image input resolutions centered about the training resolution of the model (Fig. \ref{figure:robustness}).
Hence, \modelname, while still susceptible to some spurious correlations, is far more robust than a standard vision-language model.


As points of comparison, all prior work train a supervised model on the training split. In contrast, our predictions are zero-shot, and we do not use the training set. This difference makes a direct comparison of the raw accuracy difficult. That said, the best supervised training methods achieve a domain gap $\Delta$ of 4\% to 8\% (Tab. 1 and priv. correspondence, \citet{liu2021just}), comparable to our results. We take these results to indicate that our zero-shot approach leveraging perceptual grouping provides another approach for addressing spurious correlations and learning robust image features.

\subsection{Ablation Studies}
\label{section:ablations}

We next perform experiments to demonstrate how individual factors in \modelname led to improved localization. 

We first explore the effect of pre-trained representations. In \cref{tbl:ablation1}, we freeze each of the backbones with self-supervised pre-training \cite{caron2021emerging} for the image backbone and sentence T5 pre-training \cite{ni2021sentence} for the text backbone. Our ablations indicate that the pre-trained weights alone do not contribute to the strong perceptual grouping of \modelname: our modified training process is necessary. In fact, both classification and semantic segmentation performance is affected negatively by freezing either backbone. 

We also explore how alternate or no pre-training effects overall performance. \Cref{tbl:ablation2} explores the selective removal of pre-training on the image model, language model or both. All models employ maximum pooling aggregation across spatial locations. Again, we see that \modelname exhibits significant drops in both zero-shot image recognition and localization by selectively dropping out each pre-training step. 
For instance, model performance drops from 42.3\% to 25.6\% top-1 accuracy. Likewise, the semantic segmentation mIoU drops from 50.8\% to 23.5\% accuracy. 
As expected, ImageNet supervised pre-training improves ImageNet top-1 accuracy, but interestingly leads to significant drops in grouping performance. 
For bottom-up segmentation, initializing from pretrained models benefits from scaling up the joint training data (Tab. \ref{tbl:ablation2} vs. \ref{table:ablations-internal-data-init}).
We suspect that these results indicate how each initialization provides valuable prior information not readily available in joint training for eliciting strong grouping properties, while also demonstrating the need for our training mechanism to emerge such grouping behaviour.

\begin{table}[t]
\small
\centering
\scalebox{0.92}{
\def\arraystretch{1.0}
\setlength\tabcolsep{0.9em}
\centering
\begin{tabular}{c|c|c|c|c}
\toprule
Aggregation & I-F & T-F & IN (Acc) & VOC (mIoU) \\ \midrule
Cls & \cmark & \xmark & 39.9 & 3.4 \\
Max & \cmark & \xmark & 24.2 & 10.4 \\
Max & \xmark & \cmark & 35.9 & 29.5 \\
Max & \xmark & \xmark & \textbf{42.3} & \textbf{50.8} \\ \bottomrule
\end{tabular}
}
\vspace{-0.8em}
\caption{
\textbf{Ablation on freezing pre-trained backbones:} We report Top-1 accuracy (\%) for ImageNet (IN) and mean IoU for VOC.
I-P stands for image backbone pooling, I-F stands for image backbone frozen, and T-F stands for text backbone frozen. 
}
\label{tbl:ablation1}
\end{table}

\begin{table}[t]
\small
\centering
\scalebox{0.90}{
\def\arraystretch{1.1}
\setlength\tabcolsep{1.0em}
\begin{tabular}{c|cc|ccc}
   \toprule
   \multirow{2}{*}{Dataset} & Image & T5 & ImageNet & \multicolumn{2}{c}{Pascal VOC} \\
   & Init & Init? & \scriptsize{Accuracy} & \scriptsize{mIoU} & \scriptsize{Jaccard} \\
   \shline
   \multirow{6}{*}{CC-12M} & DINO & \cmark & 42.3 & \textbf{50.8}  & \textbf{47.5} \\
   & IN-1K & \cmark & 53.3 & 22.5 & 43.3 \\
   & random & \cmark & 28.9 & 32.9  & 43.6 \\
   & DINO & & 34.1 & 44.3  & 47.2\\
   & IN-1K & & 44.5 & 20.0 & 42.2 \\
   & random & & 25.6 & 23.5  & 43.1\\
   \bottomrule
\end{tabular}}
\vspace{-0.8em}
\caption{
\textbf{Ablation on alternate pre-training:} 
We report Top-1 accuracy (\%) for ImageNet and mean IoU \& Jaccard Similarity for VOC.
Image encoder is initialized with DINO, supervised training on ImageNet-1K, or random weights. Text encoder is initialized with Sentence T5 or random weights. 
Parallel ablations using \datasetname-134M in App. \ref{appendix:ablation-studies}.
}
\label{tbl:ablation2}
\vspace{-1em}
\end{table}

We next ablate the choice of aggregation mechanism. \modelname employs a maximum operation over all spatial locations. We likewise train models performing spatial averaging or employing a class token. We present these results in \cref{tbl:ablation3}. The standard procedures of class token and average pooling result in similar performance on zero-shot classification on ImageNet, but notable reductions in mIoU on VOC semantic segmentation. For instance, in the model trained with CC-12M, mIoU on VOC drops from 50.8\% to 4.0\% representing a relative drop of 91.3\%. Similarly, in the case of bottom-up grouping on the same dataset, we demonstrate a 10 point drop in JS. We use these results to highlight the significant role played by the aggregation mechanism in inducing observed grouping properties.

We finally explore the effect of proposed token sub-sampling in \cref{tbl:ablation4}. Improvements in classification and semantic segmentation performance across datasets verify its role in boosting performance.

\begin{table}[t]

\small
\centering
\scalebox{0.90}{
\def\arraystretch{1.1}
\setlength\tabcolsep{1.1em}
\begin{tabular}{c|c|ccc}
\toprule
   \multirow{2}{*}{Dataset} & \multirow{2}{*}{Aggreg.} & ImageNet & \multicolumn{2}{c}{Pascal VOC} \\
   & & \scriptsize{Accuracy} & \scriptsize{mIoU} & \scriptsize{Jaccard} \\
   \shline
   \multirow{3}{*}{CC-12M} & Max & 42.3 & \textbf{50.8} & \textbf{47.5} \\
   & Avg & 44.0 & 11.6 & 38.1 \\
   & Cls & 46.0 & 4.0 & 40.4 \\
   \hline   
   \multirow{3}{*}{\datasetname-134M} & Max & 59.0 & \textbf{50.1} & \textbf{54.6} \\ 
   & Avg & 60.0 & 17.9 &40.5 \\ 
   & Cls & 60.2 & 4.1 & 41.3 \\  
   \bottomrule
\end{tabular}
}
\vspace{-0.8em}
\caption{
\textbf{Ablation across aggregation methods:} We report Top-1 accuracy (\%) for ImageNet and mean IoU \& Jaccard Similarity for VOC.
 Global max pooling (Max), global average pooling (Avg), and class token (Cls) alternatives are explored. All models initialized with the same pre-trained features. 
}
\label{tbl:ablation3}
\end{table}

\begin{table}[t]
\small
\centering
\scalebox{0.90}{
\def\arraystretch{1.1}
\setlength\tabcolsep{1.1em}
\begin{tabular}{c|cccc}
\toprule
TSS & IN   & VOC  & COCO & ADE20K \\ \midrule
\xmark    & 45.3 & 50.9 & 23.5 & 12.6    \\ 
\cmark    & \textbf{45.6} & \textbf{51.8} & \textbf{24.1} & \textbf{13.4}    \\  \bottomrule
\end{tabular}}
\vspace{-0.8em}
\caption{
\textbf{Ablation on token sub-sampling:} We report top-1 accuracy (\%) for ImageNet (IN) and mean IoU for the three segmentation datasets (VOC, COCO, ADE20K). TSS stands for token sub-sampling.
}
\label{tbl:ablation4}
\vspace{-1.0em}
\end{table}

\section{Discussion}
In this work we demonstrated how contrastive vision-language models have a profound lack of understanding object location. We described a minimal set of changes to existing vision-language models by modifying the aggregation method, introducing optimal pre-training strategies, and train-time token sub-sampling techniques to endow the model with both bottom-up and top-down perceptual grouping. We emphasize that our changes are minimal but sufficient to match if not exceed the performance of custom-built architectures \cite{xu2022groupvit,Xu2023LearningOS} in achieving perceptual grouping. 
We demonstrate that our resulting model provides state-of-the-art results in terms of unsupervised segmentation, and competitive results in terms of zero-shot semantic segmentation -- even though the model has been afforded \textit{no} segmentation annotations whatsoever.
Finally, we demonstrate the utility of these representations by demonstrating how perceptual grouping may be leveraged to learn visual features that are robust to spurious correlations.

We take these results to indicate that contrastive vision-language models may provide the emergence of perceptual grouping without supervision. We do see limitations in this approach 
as segmentation suffers with increasing visual clutter and label cardinality (e.g. ADE-20K). 
We suspect that recent advent of larger-scale open datasets \cite{schuhmann2021laion,kakaobrain2022coyo-700m} and advances in self-supervised learning \cite{hamilton2022unsupervised,MelasKyriazi2022DeepSM} may offer opportunities to demonstrate further benefits for endowing models with perceptual grouping. We also note the possibility of biases in our training data that may be reflected in our models. 

\vspace{0.2em}
\subsection*{Reproducibility Statement}
We built a codebase derived from OpenAI CLIP source code (\texttt{https://github.com/openai}). Source code changes to accommodate our modifications to architecture and training were minimal, and are all documented in Sec. \ref{sec:methods}. We employed pretrained \textit{Sentence T5 Base} language models from HuggingFace (\texttt{https://huggingface.co}) and image models from \cite{caron2021emerging,dosovitskiy2020image}. The CC-12M dataset was downloaded from \cite{changpinyo2021conceptual} and provides a publicly reproducible benchmark. \datasetname-134M can not be publicly released due to copyright issues. All reported numbers also contain an equivalent version for a model trained only on CC-12M to enable reproducibility.

\vspace{0.2em}
\subsection*{Ethics Statement}
We describe a minimal set of changes to vision-language models to endow these models with perceptual grouping and localization information. Our work contributes to a large literature for how to build more performant and generalizable vision models. We use both public and private computer vision datasets and leverage pretrained language and image models for our experiments. Although we believe our code and model architecture to contain no inherent bias, both the public and private data we employ may contain such biases. Any trained model should thus be approached and deployed with caution to ensure that that all fairness and bias issues are properly addressed.
\vspace{0.2em}
\subsection*{Acknowledgments}
We would like to thank the broader research team at Apple for support, feedback and guidance. In particular, we would like to thank Tom Gunter, Tom Nickson, Chen Chen, Floris Weers, Arjun Desai, Soroush Abbasi Koohpayegani, Dan Busbridge, Oncel Tuzel, Navdeep Jaitly, and Albin Madappally Jose for technical feedback and support; Samy Bengio, Barry Theobald, Russ Webb, Ayesha Rehman, and Marco Zuiliani for logistical help and encouragement throughout the project.


{\small
\bibliographystyle{ieee_fullname}
\bibliography{paper}
}

\newpage
\appendix

\twocolumn[
\centering
\vspace{1em}
\textsc{\large Appendix: \mbox{Perceptual Grouping in Contrastive Vision-Language Models}}
\vspace{3em}
]

\subsection*{Contributions}
\vspace{-0.2em}
KR led the project by performing most of experiments, evaluations and visualizations.
BM helped build the codebase, debugged many issues, worked on the evaluations, and discussed many of the pooling methods explored.
SR ran early experiments on localization datasets to explore potential applications in transfer learning and detection.
YY discussed all aspects of the project idea and scope, built and trained the initial CLIP implementation, and contributed many ideas to the pooling and initialization.
AT advised on the project, proposed many pretraining and pooling experiments, analyzed results, contributed to writing and reviewed code.
JS organized the project, set the research direction, discussed all aspects of the project idea and scope, and wrote much of the paper.


\section{Additional Quantitative Results}
\label{app:more-res}
In this section, we present additional quantitative evaluations for better understanding our experimental outcomes. We first extend our robustness analysis, followed by additional ablations on aggregation and pretraining strategies. 

\subsection{Robustness Analysis}
\label{app:more-robust}
We explore the open-source CLIP model from \cite{radford2021clip}, evaluating its performance on the waterbirds dataset for signs of spurious correlations learned by the model. These results presented in \cref{tbl:clip-waterbirds} illustrate how classification performance for the same set of foreground objects drops drastically against different conflicting backgrounds, even in spite of the task involving fine-grained categories of birds. 

\begin{table}[h]
    \centering
    \small
    \def\arraystretch{1.3}
    \setlength\tabcolsep{1.5em}
    \begin{tabular}{c|cc|c}
        \toprule
        CLIP \cite{radford2021clip} & water & land & $\Delta$ \\
        \shline
        waterbird & 88.3 & 70.1 & {\bf \textcolor{red}{-18.2}} \\
        landbird & 90.5 & 99.1 & {\bf \textcolor{red}{-8.6}}  \\
        \bottomrule
    \end{tabular}
    \caption{\small 
    {\bf Waterbirds evaluation for OpenAI CLIP model.} We demonstrate how even a CLIP model trained on significantly more data (400M image-text pairs) contains stronger sensitivty to spurious correlations than \modelname trained with an order of magnitude less data (12M). The first two columns report top-1 classification accuracy (\%) and the last column reports difference of diagonal and off-diagonal terms (delta). Higher spurious correlations increase the absolute value of delta.
    }
    \label{tbl:clip-waterbirds}
\end{table}

\subsection{Alternate Aggregation Strategies}
\label{section:additional-aggregation}
Having explored two standard visual aggregation techniques as baselines, we ask how aggregation on the textual encoder affects \modelname performance. In detail, we first explore maximum pooling for the language modality. Thereafter, we draw further attention to the visual modality, exploring more complex pooling strategies.

\textbf{Language Modality.}
In Tab. \ref{table:text-pooling}, we explore how pooling on the text embedding affects overall performance. We replace default average pooling with a maximum pooling operation to discover drops in performance across all metrics. This poor performance of maximum pooling based aggregation for language is consistent with prior works \cite{Harwath2019JointlyDV}. We hypothesize the reasoning as the nature of our task: while we attempt a localization across the visual modality, on the language end we reason with the entire text prompt as a single unit. 

\textbf{Visual Modality.}
We explored a range of alternate aggregation strategies that performed subpar to spatial max pooling utilized in \modelname. Two noteworthy approaches include Text Similarity Pooling (TSP) and Weighted Maximum Pooling (WMP). In TSP, we measure the similarity of each spatial token (corresponding to different positions) and obtain a normalized distribution using a softmax operation (including a temperature for smoothing). We aggregate the visual modality using a weighted averaging operation where the similarity of each spatial location to the text is the weight.
WMP follows the same idea but employs per-channel per-location embedding values instead of similarity as weights for the aggregation operation. 
Tab. \ref{table:alternate-pooling} shows results for TSP, WMP and max pooling.
TSP performs poorly across all variations while WMP works better at lower temperatures. Given the common softmax operation, higher temperature values result in smoother weights for both cases, making WMP and TSP more similar to average pooling. In the case of WMP, lower temperatures make the operation similar to simple spatial maximum pooling, which is reflected in the improved results for lower temperature values. 

\begin{table}[t]
\centering
\scalebox{0.90}{
\def\arraystretch{1.0}
\begin{tabular}{l|cccc}
\toprule
Method & IN   & VOC  & COCO & ADE-20K \\ \midrule
Avg    & 45.3 & 50.8 & 23.8 & 13.1    \\ 
Max    & 42.0 & 42.6 & 18.9 & 10.5    \\ \bottomrule
\end{tabular}}
\vspace{-0.5em}
\caption{\small {\bf Alternate Pooling Strategies for Text Modality.} Unlike the visual modality, performance drops when replacing the default average pooling (Avg) with maximum pooling (Max).}
\label{table:text-pooling}
\end{table}
\begin{table}[t]
\centering
\scriptsize
\centering
\scalebox{1.1}{
\def\arraystretch{1.1}
\begin{tabular}{c|cc|ccc}
   \toprule
   \multirow{2}{*}{dataset} & image & T5 & ImageNet & \multicolumn{2}{c}{Pascal VOC} \\
   & init & init? & \scriptsize{accuracy} & \scriptsize{mIoU} & \scriptsize{Jaccard} \\
   \shline
   \multirow{6}{*}{\shortstack[c]{\datasetname\\-134M}} & DINO & \cmark & 59.0 & 50.1 & 54.6\\
   & IN-1K & \cmark & 63.6 & 21.7 & 40.2 \\
   & random& \cmark & 49.4 & 37.3 & 46.3 \\
   & DINO & & 55.2 & 46.9 & 53.7 \\
   & IN-1K & & 60.6 & 20.9 & 39.1 \\
   &random & & 46.4 & 37.1 & 45.8 \\
   \bottomrule
\end{tabular}
}
\caption{\small
\textbf{Additional ablation studies with \datasetname-134M}. Parallel results for Table \ref{tbl:ablation2} (center) for ablations on weight initialization. We observe similar trends in results across these experiments.}
\label{table:ablations-internal-data-init}
\end{table}


\begin{table*}[t]
\centering
\begin{minipage}{0.49\textwidth}
\scalebox{0.9}{
\centering
\def\arraystretch{1.0}
\begin{tabular}{l|ccccc}
\toprule
Method & temp & IN      & VOC   & COCO  & ADE-20K \\ \midrule
TSP    & 0.1         & 0 & 2.98  & 0     & 0       \\
TSP    & 1.0         & 20.15   & 4.91  & 1.25  & 0       \\
TSP    & 10          & 24.70   & 15.83 & 4.27  & 2.29    \\ \midrule
Max    &             & 27.05   & 37.39 & 17.32 & 9.69    \\ \bottomrule
\end{tabular}
}
\end{minipage}
\begin{minipage}{0.49\textwidth}
\scalebox{0.9}{
\centering
\def\arraystretch{1.0}
\begin{tabular}{l|ccccc}
\toprule
Method & temp        & IN    & VOC   & COCO  & ADE-20K \\ \midrule
WMP    & 0.1         & 28.36 & 31.12 & 13.64 & 8.12    \\
WMP    & 1.0         & 0     & 0     & 0     & 0       \\
WMP    & 10          & 0     & 3.53  & 0     & 0       \\ \midrule
Max    &             & 27.05 & 37.39 & 17.32 & 9.69    \\ \bottomrule
\end{tabular}}
\end{minipage}
\caption{
{\bf Alternate Pooling Strategies for Visual Modality.} We report results for TSP and WMP with models trained for lesser steps on CC-12M with no initialization. Max refers to the spatial max operation used in \modelname. The temperature of the softmax operation for TSP and WMP is indicated by \textit{temp}.}
\label{table:alternate-pooling}
\end{table*}
\begin{figure*}[t]
\begin{center}
\vspace{-1em}
\includegraphics[width=0.02\textwidth]{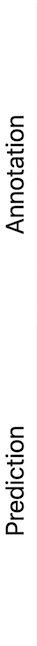} \hspace{-0.5em}
\includegraphics[width=0.97\textwidth]{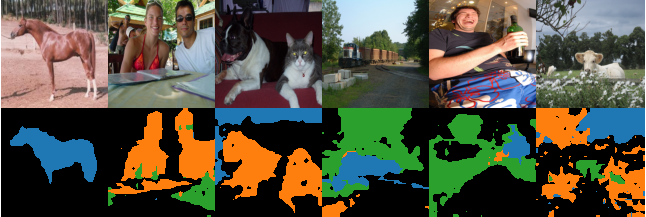}
\end{center}
\vspace{-1em}
\caption{{\bf Qualitative examples of bottom-up unsupervised segmentation with \modelname.} We illustrate examples from PASCAL VOC dataset with original image and \modelname prediction in the top and bottom rows, respectively. Note that colors correspond to clusters and {\it not} semantic labels.}
\label{fig:qualitative-bottom-up-segmentation-examples}
\end{figure*}

\subsection{Pretraining Ablations}
\label{appendix:ablation-studies}

Self-supervised pretraining of the vision head of \modelname leads to notable performance gains. We also explore how alternate supervised pretraining affects performance. In particular, we pretrain the image backbone using ImageNet-1K in a fully-supervised setting, and use the penultimate features to initialize \modelname. Apart from pretraining, all hyperparameters are unchanged. We present these results in Tab. \ref{table:ablations-internal-data-init}. 
We observe when training with \datasetname-134M that supervised pretraining leads to considerable performance gains for ImageNet-1K top-1 accuracy, but a considerable drop in segmentation performance across all three datasets. However, we note that supervised ImageNet pre-training provides unfair advantage in the case of ImageNet accuracy, in particular given the ability of overfitting those visual concepts. So we focus more on the segmentation results. 
Interestingly, entirely eliminating visual pre-training, while degrading segmentation performance, outperforms the ImageNet supervised pre-training initialized model. This reaffirms our hypothesis of better generalization of self-supervised features for segmentation tasks. We also note that results on \datasetname-134M indicate trends similar to those with CC-12M.


\section{Qualitative examples of localization}
\label{section:appendix-qualitative}

In this section, we showcase additional qualitative examples of the success and failures of \modelname on bottom-up unsupervised segmentation and top-down semantic segmentation. 

Fig. \ref{fig:qualitative-bottom-up-segmentation-examples} shows examples of bottom-up unsupervised segmentation on PASCAL VOC for \modelname. The left three examples highlight the strength of the method for images with less clutter in which there is a single object of interest. The right three examples shows examples highlights the failures in the presence of scene clutter.

\begin{figure*}[ht]
\centering
\begin{minipage}{0.72\textwidth}
\textit{Pascal VOC 2012}
\vspace{-0.5em}
\begin{center}
\includegraphics[width=0.026\textwidth]{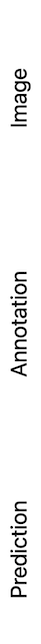} \hspace{-0.5em}
\includegraphics[width=0.81\textwidth]{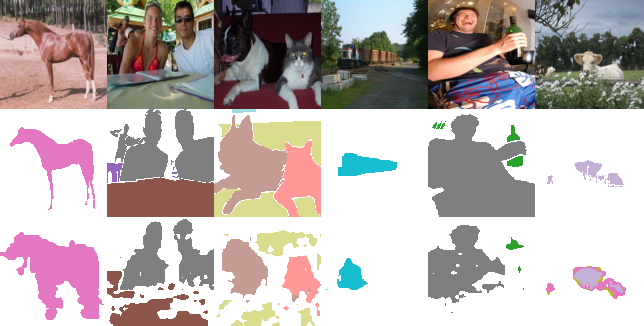} \hspace{-1em}
\includegraphics[width=0.15\textwidth]{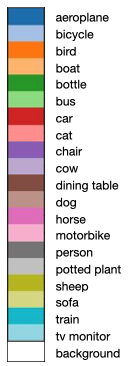}
\end{center}
\textit{COCO}
\vspace{-0.5em}
\begin{center}
\includegraphics[width=0.031\textwidth]{figures/appendix/text2.png} \hspace{-0.5em}
\includegraphics[width=0.95\textwidth]{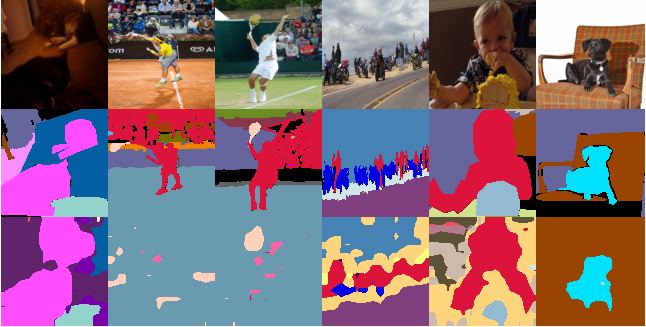}
\end{center}
\textit{ADE-20K}
\vspace{-0.5em}
\begin{center}
\includegraphics[width=0.031\textwidth]{figures/appendix/text2.png} \hspace{-0.5em}
\includegraphics[width=0.95\textwidth]{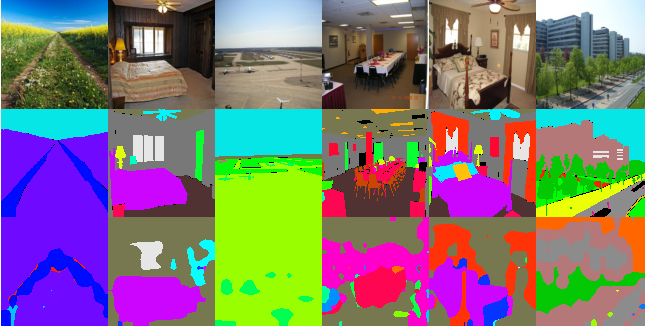} \hspace{-1em}
\end{center}
\vspace{-1em}
\caption{{\bf Qualitative examples of top-down semantic segmentation with \modelname} from PASCAL VOC, COCO and ADE-20K. For Pascal VOC, we supply a color legend for the 20 label classes. Note that the Pascal VOC examples correspond to the same examples from Fig. \ref{fig:qualitative-bottom-up-segmentation-examples}.}
\label{fig:qualitative-top-down-segmentation-examples}
\end{minipage}
\end{figure*}

Fig. \ref{fig:qualitative-top-down-segmentation-examples} present additional top-down semantic segmentation examples across all three datasets used for evaluation. The examples from PASCAL VOC dataset are the same examples from Fig. \ref{fig:qualitative-bottom-up-segmentation-examples}. For PASCAL VOC, note that the contrast between the top-down and bottom-up segmentations. The top-down segmentation is able to correctly separate the dog and cat classes (column 3) and also improve performance in the more cluttered scenes. On the other hand, in column 4, it missed out on a portion of the train that was segmented properly in the bottom-up setting. Segmentations from \modelname may contain discontinuities within a single object region in some cases, especially for the background objects (columns 2-3).

COCO and ADE-20K provide more challenging datasets and highlight several potential failure modes. A notable failure mode for \modelname is to reverting to the baseline CLIP behaviour by predicting the salient object class at all locations. This is visible to some extent in column 2 \& 3 in the COCO examples and column 3 in the ADE-20K examples. Additionally, \modelname results in false positives for cluttered scenes as visible in some examples of these two datasets. 

In summary, \modelname is able to localize the salient objects well in less cluttered scenes, even when multiple objects belonging to different classes are present. \modelname is also able to coarsely localize some of the salient objects in more cluttered scenes. In terms of limitations, \modelname fails to correctly localize some background classes, fails to correctly recognize object boundaries, and may miss smaller objects in cluttered scenes.

\section{Details of \datasetname-134M dataset}
\label{section:internal-dataset}
High Quality Image Text Pairs (\datasetname-134M) consists of $\sim$134 million diverse and high quality images paired with descriptive captions and titles. Images range in spatial resolution from 320 to 2048 pixels on the short side. All images are JPEG format and most are RGB. Each example image is associated with a title, and a list of several captions. A small fraction (<< 1\%) of the examples are missing both captions and title. We favor the associated captions, and find that these tokenize to an average length of 20.1 tokens, although the shortest caption is only one token and the longest is over 1000. This dataset was licensed to our research lab by a third party for commercial use.

To preprocess this dataset for training, we first exclude all pairs for which no valid caption exists.  We also perform global exact-byte-match image de-duplication across our full training corpus, meaning that valid examples may be dropped due to appearing in other subsets of our overall training dataset. As we draw each example, we create an image-text pair by sampling from the list of available captions.  The text is then tokenized, and the image is resized so that the shortest side is 224 pixels, with a further random crop then applied over the longer dimension to produce a 224 x 224 pixel square image. Lastly, we normalize the image using statistics derived from our full training corpus.

\section{Architecture and Training Details}
\label{section:training-details}
Our implementations of CLIP and ALIGN employ ViT-B/16 \cite{dosovitskiy2020image}, and EfficientNet-B5\cite{tan2019efficientnet} for the image embedding, respectively, to mirror the primary results presented in each respective vision-language model. For the ViT architecture, we experimented with varying patch sizes $P=8$ and $P=16$ in order to leverage open-sourced DINO pretrained weights \cite{caron2021emerging}, but report all of our results with $P=16$. 

We train all models on 224$\times$224 images to provide a fair comparison with \citet{radford2021clip}. Note however that the published version of ALIGN employed a 640$\times$640 resolution. ViT models may operate on images at arbitrary spatial resolution. At inference time we experimented with spatial resolutions of 224$\times$224 and 448$\times$448, resulting in 196 and 784 tokens, respectively. Results were similar across 224 and 448 resolutions, we report only results at 224 for brevity (except for Fig. \ref{figure:robustness}). 

We train models with 32 GPUs across 4 machines with PyTorch \cite{NEURIPS2019_bdbca288} using the LAMB optimizer \cite{you2019large} with a cosine decayed learning rate with linear warm-up. We employ an initial learning rate of 1e-3, 2000 warm-up steps, and decay the rate with a single period over the entire training regime (32 epochs for CC12M; 10 epochs for \datasetname-134M). We employ a weight decay of 1e-2. All training parameters were determined through moderate hyperparameter tuning.

\textbf{Patch Sub-Sampling:}
During training, we also utilize overlapping patch generation with patch sub-sampling as regularization. In particular, the token sequence length is always maintained at the original value of 224 during training through random sub-sampling (patches selected according to uniform distribution). This also allows obtaining a higher resolution feature map from a fixed resolution image during inference.

\section{Limitations of CNN-based architectures}
\label{section:cnn-analysis}

The primary focus of our presentation is on the Vision Transformer (ViT) architecture \cite{dosovitskiy2020image}. The reason for this focus is that the Transformer architecture is particularly well suited for multimodal learning tasks because one does not need to craft an architecture for each modality, and tune the training set up for each particular architecture.
We also recognize that convolutional neural networks (CNN's) have a long history of providing state-of-the-art CNN results on computer vision related problems. EfficientNet-B5 is a modern state-of-the-art architecture whose meta-architecture and scaling properties were derived from architecture search considerations \cite{tan2019efficientnet} (but see \citet{bello2021revisiting}), and provides the visual backbone for the ALIGN model \cite{jia2021scaling}.

We experimented with this model and find that the image featurization from a CNN-derived backbone achieves favorable results with respect to previously published ViT models on localization problems. Table \ref{table:semantic-segmentation} showcases higher mIoU for semantic segmentation than a model with a ViT backbone (CLIP) on ADE20K, COCO and Pascal VOC when trained on the same dataset. Likewise, previous results published on ALIGN with an EfficientNet-B5 backbone achieved superior results to a ViT model on COCO and ADE20K in terms of semantic segmentation \footnote{We note that the previously published ALIGN results were based on a backbone trained with a much larger dataset (1.8B image-text pairs), and it was evaluated at a higher resolution of 640$\times$640 pixels, resulting in a zero-shot image recognition performance of 76.4. In comparison, our baseline and proposed models operate at 224$\times$224 resolution and was trained on a 10$\times$ smaller dataset.}.

Most importantly, in spite of many attempts, we were not able to improve the performance of the EfficientNet-B5 architecture for localization. The best results for a CNN-based architecture we achieved are shown in Table \ref{table:semantic-segmentation}, which are notably below previously published results and our best results with a ViT architecture. At best, the addition of maximum pooling at the top layer of a CNN led to marginal gains in terms of mIoU or Jaccard similarity. We suspect that a custom architecture (such as ASPP or FPN) may improve these results further \cite{chen2017deeplab,lin2017feature}, but we consider this out of scope as we are attempting to learn a featurization that does not artificially increase the parameterization in order to solve a specialized task of localization. We suspect that this limitation of CNN architectures may reflect the fact that CNN architecture already have learned a representation that is heavily dependent on the spatial geometry derived from a convolutional kernel. Such an inductive bias may be not provide a suitable mechanism for providing global processing in a segmentation task \cite{wang2018non}.

\begin{table*}[ht]
\centering
\begin{minipage}{0.70\textwidth}
\scalebox{0.95}{
\def\arraystretch{1.0}
\begin{tabular}{@{ }l|cc|ccc}
\toprule
   & \small{segment} & \small{segment} & & & \\
   & \small{label?} & \small{mask?} & \small{ADE20K} & \small{COCO} & \small{PASCAL VOC} \\
   \shline
   SPNet \citet{xian2019semantic} & \cmark & \cmark &  & & 18.3 \\
   ZS3Net \citet{bucher2019zeroshotseg} & \cmark & \cmark &  &  & 38.3 \\   
   LSeg \citet{li2022language} & \cmark & \cmark &  & 27.2 & 47.4 \\
   LSeg+ \citet{ghiasi2021simple} & \cmark & \cmark & 18.0 & \,\,\,\,55.1 $^\dagger$ & 59.0 \\
   ALIGN w/ proposal \citet{ghiasi2021simple} & & \cmark & 12.9 & \,\,\,\,17.9 $^\dagger$ & 22.4 \\   
   OpenSeg \citet{ghiasi2021simple} & & \cmark & 21.1 & \,\,\,\,36.1 $^\dagger$ & 70.2 \\
   OpenSeg + Narr. \citet{ghiasi2021simple} & & \cmark & 24.8 & \,\,\,\,38.1 $^\dagger$ & 72.2 \\
   \bottomrule
\end{tabular}
}
\end{minipage}
\begin{minipage}{0.28\textwidth}
\caption{ \textbf{Performance of prior zero-shot segmentation models trained on segmentation data.} All numbers report the mIoU for semantic segmentation on ADE20K (150 labels), PASCAL-VOC (20 labels) and COCO (50 labels). $\dagger$ indicates models that were trained on image segmentation masks from the COCO dataset.}
\label{table:semantic-segmentation-annotations}  
\end{minipage}
\end{table*}

\section{Prior work on zero shot semantic segmentation}
\label{section:prior-zemantic-segmentation}

Recent work has made impressive strides on zero-shot semantic segmentation. These works focus on training such models on subsets of segmentation data, whether masks and/or labels and test the performance of the resulting model on other splits of data. The zero-shot performance is assessed by splitting the labeled datasets to ensure that the model is tested in a zero-shot manner on unseen labels.  Table \ref{table:semantic-segmentation-annotations} summarizes results from several recent papers \cite{ghiasi2022open,li2022language,xian2019semantic,bucher2019zeroshotseg}

We note that particularly later forms of models achieve results superior to those presented here \cite{ghiasi2022open,li2022language}, but we emphasize several important distinctions. First, these models were trained on segmentation masks in order to learn perceptual grouping across visual imagery. Our work instead addresses how a model may learn this information without being explicitly supplied examples teaching such behavior. Second, most of the models were trained using segmentation masks from the COCO dataset. Hence, these models might perform particularly well on this dataset making comparisons on COCO less comparable to our model.

\section{Prior work on unsupervised segmentation}
\label{section:prior-unsupervised-segmentation}
The task of unsupervised segmentation groups semantically related concepts using only pixel information. 
Recent advances in self-supervised learning (e.g. \citet{caron2021emerging}) has led to numerous opportunities across tasks \cite{Li2022DoesSL,Shang2021SelfSupervisedDR}, including for unsupervised segmentation or weakly-supervised localization tasks \cite{hamilton2022unsupervised, cho2021picie, VanGansbeke2021UnsupervisedSS, Ji2019InvariantIC,Shang2022LearningVV}. These methods train and operate with no semantic labels, performing bottom-up grouping of image content.
A common characteristic is the presence of iterative clustering mechanisms (e.g. $k$-Means clustering \cite{MacQueen1967SomeMF}). Notably, STEGO \cite{hamilton2022unsupervised} employs $k$-Means and Conditional Random Fields (CRF) \cite{Krhenbhl2011EfficientII} to refine and improve the inference procedure. We hope to apply CRFs (e.g. \cite{Jayasumana2019BipartiteCR}) to \modelname in future.
While \modelname exhibits competitive performance, we note that techniques proposed by other methods (e.g. \cite{hamilton2022unsupervised}) may be leveraged (and combined with \modelname) to refine our segmentation performance. A key distinction of our work is the focus on learning a new representation space aligned to language that exhibits strong perceptual grouping.

\section{Details of robustness analysis}
\label{section:waterbirds-details}

We calculate the zero-shot prediction of the class in a non-standard manner to exploit the spatial reasoning of \modelname. We apply the same zero-shot evaluation to the baseline CLIP model. Specifically, we first calculate the embedding for all labels within each category of \texttt{waterbird}, \texttt{landbird} and \texttt{background}. For each of these categories we calculate the average image embeddings across these labels at each spatial location.

To exploit the spatial knowledge of the model, we focus our analysis on all spatial locations which are \textit{not} labeled as \texttt{background}. For all locations which maximally predict a \texttt{waterbird}, we calculate the similarity to the embedding for a \texttt{waterbird}. Likewise, we do the same for all locations maximized by \texttt{landbird}. Finally, our resulting prediction is the class that is closest to its associated embedding.

We find that these results and the corresponding robustness vary substantially due to the selection of prompts for each of the three categories. This matches observations in \citet{radford2021clip}. In Sec. \ref{sec:robustness} we focus on the results of the model trained on CC-12M using the prompts listed below which contain a minimal amount of prompt engineering. In particular, the prompts for \texttt{waterbirds} and \texttt{landbirds} follow \citet{sagawa2019distributionally}. See also \citet{WelinderEtal2010}.
\begin{itemize}[leftmargin=*]
\small
\item \texttt{background}: background
\item \texttt{waterbird}: Black footed Albatross, Laysan Albatross, Sooty Albatross, Crested Auklet, Least Auklet, Parakeet Auklet, Rhinoceros Auklet, Brandt Cormorant, Red faced Cormorant, Pelagic Cormorant, Frigatebird, Northern Fulmar, Gadwall, Eared Grebe, Horned Grebe, Pied billed Grebe, Western Grebe, Pigeon Guillemot, California Gull, Glaucous winged Gull, Heermann Gull, Herring Gull, Ivory Gull, Ring billed Gull, Slaty backed Gull, Western Gull, Long tailed Jaeger, Pomarine Jaeger, Red legged Kittiwake, Pacific Loon, Mallard, Hooded Merganser, Red breasted Merganser, Brown Pelican, White Pelican, Horned Puffin, Artic Tern, Black Tern, Caspian Tern, Common Tern, Elegant Tern, Forsters Tern, Least Tern
\item \texttt{landbird}: Groove billed Ani, Brewer Blackbird, Red winged Blackbird, Rusty Blackbird, Yellow headed Blackbird, Bobolink, Indigo Bunting, Lazuli Bunting, Painted Bunting, Cardinal, Spotted Catbird, Gray Catbird, Yellow breasted Chat, Eastern Towhee, Chuck will Widow, Bronzed Cowbird, Shiny Cowbird, Brown Creeper, American Crow, Fish Crow, Black billed Cuckoo, Mangrove Cuckoo, Yellow billed Cuckoo, Gray crowned Rosy Finch, Purple Finch, Northern Flicker, Acadian Flycatcher, Great Crested Flycatcher, Least Flycatcher, Olive sided Flycatcher, Scissor tailed Flycatcher, Vermilion Flycatcher, Yellow bellied Flycatcher, American Goldfinch, European Goldfinch, Boat tailed Grackle, Blue Grosbeak, Evening Grosbeak, Pine Grosbeak, Rose breasted Grosbeak, Anna Hummingbird, Ruby throated Hummingbird, Rufous Hummingbird, Green Violetear, Blue Jay, Florida Jay, Green Jay, Dark eyed Junco, Tropical Kingbird, Gray Kingbird, Belted Kingfisher, Green Kingfisher, Pied Kingfisher, Ringed Kingfisher, White breasted Kingfisher, Horned Lark, Western Meadowlark, Mockingbird, Nighthawk, Clark Nutcracker, White breasted Nuthatch, Baltimore Oriole, Hooded Oriole, Orchard Oriole, Scott Oriole, Ovenbird, Western Wood Pewee, Sayornis, American Pipit, Whip poor Will, Common Raven, White necked Raven, American Redstart, Geococcyx, Loggerhead Shrike, Great Grey Shrike, Baird Sparrow, Black throated Sparrow, Brewer Sparrow, Chipping Sparrow, Clay colored Sparrow, House Sparrow, Field Sparrow, Fox Sparrow, Grasshopper Sparrow, Harris Sparrow, Henslow Sparrow, Le Conte Sparrow, Lincoln Sparrow, Nelson Sharp tailed Sparrow, Savannah Sparrow, Seaside Sparrow, Song Sparrow, Tree Sparrow, Vesper Sparrow, White crowned Sparrow, White throated Sparrow, Cape Glossy Starling, Bank Swallow, Barn Swallow, Cliff Swallow, Tree Swallow, Scarlet Tanager, Summer Tanager, Green tailed Towhee, Brown Thrasher, Sage Thrasher, Black capped Vireo, Blue headed Vireo, Philadelphia Vireo, Red eyed Vireo, Warbling Vireo, White eyed Vireo, Yellow throated Vireo, Bay breasted Warbler, Black and white Warbler, Black throated Blue Warbler, Blue winged Warbler, Canada Warbler, Cape May Warbler, Cerulean Warbler, Chestnut sided Warbler, Golden winged Warbler, Hooded Warbler, Kentucky Warbler, Magnolia Warbler, Mourning Warbler, Myrtle Warbler, Nashville Warbler, Orange crowned Warbler, Palm Warbler, Pine Warbler, Prairie Warbler, Prothonotary Warbler, Swainson Warbler, Tennessee Warbler, Wilson Warbler, Worm eating Warbler, Yellow Warbler, Northern Waterthrush, Louisiana Waterthrush, Bohemian Waxwing, Cedar Waxwing, American Three toed Woodpecker, Pileated Woodpecker, Red bellied Woodpecker, Red cockaded Woodpecker, Red headed Woodpecker, Downy Woodpecker, Bewick Wren, Cactus Wren, Carolina Wren, House Wren, Marsh Wren, Rock Wren, Winter Wren, Common Yellowthroat
\end{itemize}

\section{Prompts for Zero-shot Segmentation}
\label{section:prompts}

We employed the following prompts for probing our vision-language models for zero-shot semantic segmentation. These prompts were copied from the corresponding label sets of each dataset with some basic considerations, for instance, restoring spaces in compound words. For prompts separated by a comma, the average embedding across all prompts delineated by commas is associated with each label.

\vspace{-0.2cm}
\paragraph{\textbf{Pascal VOC 2012} \cite{everingham2010pascal}}

\begin{enumerate}[leftmargin=*]
\small
	\item \texttt{background}: background, crops, bush, shrub, tiles, pavement, rug, carpet, box, boxes, speaker, storage, painting, board, panel, poster, clock, cage, drinking glass, park, plaything, toy, fireplace, bag, bag, bed, bench, book, books, building, buildings, cabinet, drawer, ceiling, computer, computer case, cup, cups, door, fence, floor, flower, grass, lawn, turf, ground, soil, dirt, tiles, keyboard, lamp, mountain, hills, mouse, curtain, platform, sign, street, rock, stone, shelf, sidewalk, sky, clouds, snow, track, train track, tree, trees, wall, water, window, wood, woods
	\item \texttt{aeroplane}: aeroplane, airplane, aeroplanes, airplanes
	\item \texttt{bicycle}: bicycle, bicycles, bike, bikes
	\item \texttt{bird}: bird, birds
	\item \texttt{boat}: boat, boats
	\item \texttt{bottle}: bottle, bottles, water bottle
	\item \texttt{bus}: bus, buses
	\item \texttt{car}: car, cars
	\item \texttt{cat}: cat, cats, kitties, kitty
	\item \texttt{chair}: chair, chairs
	\item \texttt{cow}: cow, cows, calf
	\item \texttt{diningtable}: diningtable, dining table, diningtables, dining tables, plate, plates
	\item \texttt{dog}: dog, dogs, puppy, puppies
	\item \texttt{horse}: horse, horses, foal
	\item \texttt{motorbike}: motorbike, motorcycle, motorbikes, motorcycles
	\item \texttt{person}: person, child, girl, boy, woman, man, people, children, girls, boys, women, men, lady, guy, ladies, guys, clothes
	\item \texttt{pottedplant}: pottedplant, pottedplants, plant pot, plant pots, planter, planters, potted plant
	\item \texttt{sheep}: sheep
	\item \texttt{sofa}: sofa, sofas
	\item \texttt{train}: train, trains, locomotive, locomotives, freight train
	\item \texttt{tvmonitor}: tvmonitor, monitor, tv, televison, television monitor
    \end{enumerate}

\vspace{-0.2cm}
\paragraph{\textbf{COCO 2017} \cite{coco}} 
\small
\begin{enumerate}[leftmargin=*]
\item \texttt{airplane}: 	airplane
\item \texttt{apple}: 	apple
\item \texttt{backpack}: 	backpack
\item \texttt{banana}: 	banana
\item \texttt{baseball bat}: 	baseball bat
\item \texttt{baseball glove}: 	baseball glove
\item \texttt{bear}: 	bear
\item \texttt{bed}: 	bed
\item \texttt{bench}: 	bench
\item \texttt{bicycle}: 	bicycle
\item \texttt{bird}: 	bird
\item \texttt{boat}: 	boat
\item \texttt{book}: 	book
\item \texttt{bottle}: 	bottle
\item \texttt{bowl}: 	bowl
\item \texttt{broccoli}: 	broccoli
\item \texttt{bus}: 	bus
\item \texttt{cake}: 	cake
\item \texttt{car}: 	car
\item \texttt{carrot}: 	carrot
\item \texttt{cat}: 	cat
\item \texttt{cell phone}: 	cell phone
\item \texttt{chair}: 	chair
\item \texttt{clock}: 	clock
\item \texttt{couch}: 	couch
\item \texttt{cow}: 	cow
\item \texttt{cup}: 	cup
\item \texttt{dining table}: 	dining table
\item \texttt{dog}: 	dog
\item \texttt{donut}: 	donut
\item \texttt{elephant}: 	elephant
\item \texttt{fire hydrant}: 	fire hydrant
\item \texttt{fork}: 	fork
\item \texttt{frisbee}: 	frisbee
\item \texttt{giraffe}: 	giraffe
\item \texttt{hair drier}: 	hair drier
\item \texttt{handbag}: 	handbag
\item \texttt{horse}: 	horse
\item \texttt{hot dog}: 	hot dog
\item \texttt{keyboard}: 	keyboard
\item \texttt{kite}: 	kite
\item \texttt{knife}: 	knife
\item \texttt{laptop}: 	laptop
\item \texttt{microwave}: 	microwave
\item \texttt{motorcycle}: 	motorcycle
\item \texttt{mouse}: 	mouse
\item \texttt{orange}: 	orange
\item \texttt{oven}: 	oven
\item \texttt{parking meter}: 	parking meter
\item \texttt{person}: 	person
\item \texttt{pizza}: 	pizza
\item \texttt{potted plant}: 	potted plant
\item \texttt{refrigerator}: 	refrigerator
\item \texttt{remote}: 	remote
\item \texttt{sandwich}: 	sandwich
\item \texttt{scissors}: 	scissors
\item \texttt{sheep}: 	sheep
\item \texttt{sink}: 	sink
\item \texttt{skateboard}: 	skateboard
\item \texttt{skis}: 	skis
\item \texttt{snowboard}: 	snowboard
\item \texttt{spoon}: 	spoon
\item \texttt{sports ball}: 	sports ball
\item \texttt{stop sign}: 	stop sign
\item \texttt{suitcase}: 	suitcase
\item \texttt{surfboard}: 	surfboard
\item \texttt{teddy bear}: 	teddy bear
\item \texttt{tennis racket}: 	tennis racket
\item \texttt{tie}: 	tie
\item \texttt{toaster}: 	toaster
\item \texttt{toilet}: 	toilet
\item \texttt{toothbrush}: 	toothbrush
\item \texttt{traffic light}: 	traffic light
\item \texttt{train}: 	train
\item \texttt{truck}: 	truck
\item \texttt{tv}: 	tv
\item \texttt{umbrella}: 	umbrella
\item \texttt{vase}: 	vase
\item \texttt{wine glass}: 	wine glass
\item \texttt{zebra}: 	zebra     
\end{enumerate}

\vspace{-0.2cm}
\paragraph{\textbf{ADE-20K} (150 frequent labels) \cite{zhou2018ade}}
\small
\begin{enumerate}[leftmargin=*]
\item \texttt{airplane}: 	airplane, aeroplane, plane
\item \texttt{animal}: 	animal, animate, being, beast, brute, creature, fauna
\item \texttt{apparel}: 	apparel, wearing, apparel, dress, clothes
\item \texttt{arcade}: 	arcade, machine
\item \texttt{armchair}: 	armchair
\item \texttt{ashcan}: 	ashcan, trash, can, garbage, can, wastebin, ash, bin, ash-bin, ashbin, dustbin, trash, barrel, trash, bin
\item \texttt{awning}: 	awning, sunshade, sunblind
\item \texttt{bag}: 	bag
\item \texttt{ball}: 	ball
\item \texttt{bannister}: 	bannister, banister, balustrade, balusters, handrail
\item \texttt{bar}: 	bar
\item \texttt{barrel}: 	barrel, cask
\item \texttt{base}: 	base, pedestal, stand
\item \texttt{basket}: 	basket, handbasket
\item \texttt{bathtub}: 	bathtub, bathing, tub, bath, tub
\item \texttt{bed}: 	bed
\item \texttt{bench}: 	bench
\item \texttt{bicycle}: 	bicycle, bike, wheel, cycle
\item \texttt{blanket}: 	blanket, cover
\item \texttt{blind}: 	blind, screen
\item \texttt{boat}: 	boat
\item \texttt{book}: 	book
\item \texttt{bookcase}: 	bookcase
\item \texttt{booth}: 	booth, cubicle, stall, kiosk
\item \texttt{bottle}: 	bottle
\item \texttt{box}: 	box
\item \texttt{bridge}: 	bridge, span
\item \texttt{buffet}: 	buffet, counter, sideboard
\item \texttt{building}: 	building, edifice
\item \texttt{bulletin}: 	bulletin, board, notice, board
\item \texttt{bus}: 	bus, autobus, coach, charabanc, double-decker, jitney, motorbus, motorcoach, omnibus, passenger, vehicle
\item \texttt{cabinet}: 	cabinet
\item \texttt{canopy}: 	canopy
\item \texttt{car}: 	car, auto, automobile, machine, motorcar
\item \texttt{case}: 	case, display, case, showcase, vitrine
\item \texttt{ceiling}: 	ceiling
\item \texttt{chair}: 	chair
\item \texttt{chandelier}: 	chandelier, pendant, pendent
\item \texttt{chest}: 	chest of drawers, chest, bureau, dresser
\item \texttt{clock}: 	clock
\item \texttt{coffee}: 	coffee, table, cocktail, table
\item \texttt{column}: 	column, pillar
\item \texttt{computer}: 	computer, computing, machine, computing, device, data, processor, electronic, computer, information, processing, system
\item \texttt{conveyer}: 	conveyer, belt, conveyor, belt, conveyer, conveyor, transporter
\item \texttt{counter}: 	counter
\item \texttt{countertop}: 	countertop
\item \texttt{cradle}: 	cradle
\item \texttt{crt}: 	crt, screen
\item \texttt{curtain}: 	curtain, drape, drapery, mantle, pall
\item \texttt{cushion}: 	cushion
\item \texttt{desk}: 	desk
\item \texttt{dirt}: 	dirt, track
\item \texttt{dishwasher}: 	dishwasher, dish, washer, dishwashing, machine
\item \texttt{door}: 	door, double, door
\item \texttt{earth}: 	earth, ground
\item \texttt{escalator}: 	escalator, moving, staircase, moving, stairway
\item \texttt{fan}: 	fan
\item \texttt{fence}: 	fence, fencing
\item \texttt{field}: 	field
\item \texttt{fireplace}: 	fireplace, hearth, open, fireplace
\item \texttt{flag}: 	flag
\item \texttt{floor}: 	floor, flooring
\item \texttt{flower}: 	flower
\item \texttt{food}: 	food, solid, food
\item \texttt{fountain}: 	fountain
\item \texttt{glass}: 	glass, drinking, glass
\item \texttt{grandstand}: 	grandstand, covered, stand
\item \texttt{grass}: 	grass
\item \texttt{hill}: 	hill
\item \texttt{hood}: 	hood, exhaust, hood
\item \texttt{house}: 	house
\item \texttt{hovel}: 	hovel, hut, hutch, shack, shanty
\item \texttt{kitchen}: 	kitchen, island
\item \texttt{lake}: 	lake
\item \texttt{lamp}: 	lamp
\item \texttt{land}: 	land, ground, soil
\item \texttt{light}: 	light, light, source
\item \texttt{microwave}: 	microwave, microwave, oven
\item \texttt{minibike}: 	minibike, motorbike
\item \texttt{mirror}: 	mirror
\item \texttt{monitor}: 	monitor, monitoring, device
\item \texttt{mountain}: 	mountain, mount
\item \texttt{ottoman}: 	ottoman, pouf, pouffe, puff, hassock
\item \texttt{oven}: 	oven
\item \texttt{painting}: 	painting, picture
\item \texttt{palm}: 	palm, palm, tree
\item \texttt{path}: 	path
\item \texttt{person}: 	person, individual, someone, somebody, mortal, soul
\item \texttt{pier}: 	pier, wharf, wharfage, dock
\item \texttt{pillow}: 	pillow
\item \texttt{plant}: 	plant, flora, plant, life
\item \texttt{plate}: 	plate
\item \texttt{plaything}: 	plaything, toy
\item \texttt{pole}: 	pole
\item \texttt{pool}: 	pool, table, billiard, table, snooker, table
\item \texttt{poster}: 	poster, posting, placard, notice, bill, card
\item \texttt{pot}: 	pot, flowerpot
\item \texttt{radiator}: 	radiator
\item \texttt{railing}: 	railing, rail
\item \texttt{refrigerator}: 	refrigerator, icebox
\item \texttt{river}: 	river
\item \texttt{road}: 	road, route
\item \texttt{rock}: 	rock, stone
\item \texttt{rug}: 	rug, carpet, carpeting
\item \texttt{runway}: 	runway
\item \texttt{sand}: 	sand
\item \texttt{sconce}: 	sconce
\item \texttt{screen}: 	screen, door, screen
\item \texttt{screen}: 	screen, silver, screen, projection, screen
\item \texttt{sculpture}: 	sculpture
\item \texttt{sea}: 	sea
\item \texttt{seat}: 	seat
\item \texttt{shelf}: 	shelf
\item \texttt{ship}: 	ship
\item \texttt{shower}: 	shower
\item \texttt{sidewalk}: 	sidewalk, pavement
\item \texttt{signboard}: 	signboard, sign
\item \texttt{sink}: 	sink
\item \texttt{sky}: 	sky
\item \texttt{skyscraper}: 	skyscraper
\item \texttt{sofa}: 	sofa, couch, lounge
\item \texttt{stage}: 	stage
\item \texttt{stairs}: 	stairs, steps
\item \texttt{stairway}: 	stairway, staircase
\item \texttt{step}: 	step, stair
\item \texttt{stool}: 	stool
\item \texttt{stove}: 	stove, kitchen, stove, range, kitchen, range, cooking, stove
\item \texttt{streetlight}: 	streetlight, street, lamp
\item \texttt{swimming}: 	swimming, pool, swimming, bath, natatorium
\item \texttt{swivel}: 	swivel, chair
\item \texttt{table}: 	table
\item \texttt{tank}: 	tank, storage, tank
\item \texttt{television}: 	television, television, receiver, television, set, tv, tv, set, idiot, box, boob, tube, telly, goggle, box
\item \texttt{tent}: 	tent, collapsible, shelter
\item \texttt{toilet}: 	toilet, can, commode, crapper, pot, potty, stool, throne
\item \texttt{towel}: 	towel
\item \texttt{tower}: 	tower
\item \texttt{trade}: 	trade, name, brand, name, brand, marque
\item \texttt{traffic}: 	traffic, light, traffic, signal, stoplight
\item \texttt{tray}: 	tray
\item \texttt{tree}: 	tree
\item \texttt{truck}: 	truck, motortruck
\item \texttt{van}: 	van
\item \texttt{vase}: 	vase
\item \texttt{wall}: 	wall
\item \texttt{wardrobe}: 	wardrobe, closet, press
\item \texttt{washer}: 	washer, automatic, washer, washing, machine
\item \texttt{water}: 	water
\item \texttt{waterfall}: 	waterfall, falls
\item \texttt{windowpane}: 	windowpane, window
\end{enumerate}

\end{document}